\documentclass[sigplan,screen]{acmart}

\copyrightyear{2026}
\acmYear{2026}
\setcopyright{cc}
\setcctype{by}
\acmConference[ASPLOS '26] {Proceedings of the 30th ACM International Conference on Architectural Support for Programming Languages and Operating Systems, Volume 2}{March 21--26, 2026}{Pittsburgh, PA, USA.}
\acmBooktitle{Proceedings of the 30th ACM International Conference on Architectural Support for Programming Languages and Operating Systems, Volume 2 (ASPLOS '26), March 21--26, 2026, Pittsburgh, PA, USA}
\acmISBN{979-8-4007-2359-9/2026/03}
\acmDOI{10.1145/3779212.3790165}

\usepackage{hyperref}
\usepackage{cleveref}
\usepackage{listings}
\usepackage{tabularx}
\usepackage{booktabs}
\usepackage{enumitem}
\usepackage{multirow}
\usepackage[frozencache, cachedir=.]{minted}
\usepackage{balance}
\usepackage{flushend}
\newcounter{packednmbr}

\usepackage{caption}
\usepackage{subcaption}
\usepackage[dvipsnames]{xcolor}
\usepackage{algorithm}
\usepackage[noend]{algpseudocode}
\usepackage{duckuments}
\usepackage{amsthm}
\usepackage{multicol}
\usepackage{multirow}
\usepackage{makecell}
\usepackage{svg}
\usepackage{graphicx}
\usepackage{float}
\usepackage{tikz}
\usetikzlibrary{decorations.pathreplacing}
\usetikzlibrary{arrows.meta}
\usetikzlibrary{fit, shapes.geometric}
\usepackage{bm}
\usepackage[normalem]{ulem} 
\usetikzlibrary{arrows.meta, positioning, calc, matrix}
\tikzset{
    block/.style = {rectangle, draw, fill=blue!20, text centered, rounded corners, minimum height=1cm},
    line/.style = {draw, -latex'}
}
\usepackage[table]{xcolor}
\definecolor{ltgray}{gray}{0.94}
\newcolumntype{G}{>{\columncolor{ltgray}}c}

\usepackage{pifont} % provides \ding{}
\newcommand{\yes}{\textcolor{green}{\ding{51}}} % checkmark
\newcommand{\no}{\textcolor{red}{\ding{55}}}    % cross mark

\makeatletter
\def\th@definition{%
  \thm@notefont{}
}
\makeatother

\theoremstyle{plain}

\theoremstyle{definition}

\algnewcommand{\LineComment}[1]{\State // #1}

\newcommand{\todoswitch}[1]{#1}

\newcommand{\todo}[1]{{\todoswitch{\color{red}{[}\textbf{TODO: #1}{]}}}}
\newcommand{\cell}[1]{\fboxrule=.1em\fboxsep=.1em\fbox{\small#1}}
\renewcommand{\cell}[1]{%
   \ensuremath{\bm{\langle}\!#1\!\bm{\rangle}}}

\colorlet{lightblue}{cyan!20}

\newcommand{\highlight}[1]{%
  \tikz[baseline=(X.base)] \node[rectangle, fill=lightblue, rounded corners, inner sep=0.3mm] (X) {#1};%
}

\newcommand{\revision}[1]{\textcolor{black}{#1}}
\newcommand{\revisionfig}[1]{
#1
}

\tikzset{
  bbox/.style={
    execute at end picture={
      \draw[blue, thick]
       ([shift={(-2pt,-2pt)}]current bounding box.south west) rectangle
        ([shift={(2pt,2pt)}]current bounding box.north east);
    }
  }
}

\newcommand{\compiler}[0]{FuseFlow}
\newcommand{\factored}[0]{factored}
\newcommand{\Factored}[0]{Factored}
\newcommand{\simulator}[0]{Comal}
\newcommand{\cycleacc}[0]{cycle-accurate}
\newcommand{\Fusiontab}[0]{Fusion table}
\newcommand{\fusiontab}[0]{fusion table}
\newcommand{\tab}[0]{table}

\begin{document}

\title[\compiler{}: A Fusion-Centric Compilation Framework for Sparse Deep Learning on Dataflow]{\compiler{}: A Fusion-Centric Compilation Framework for Sparse Deep Learning on Streaming Dataflow}

\author{Rubens Lacouture}
\affiliation{
  \institution{Stanford University}
  \city{Stanford}
  \country{USA}
}
\email{rubensl@stanford.edu}

\author{Nathan Zhang}
\affiliation{
  \institution{SambaNova Systems, Inc.}
  \city{Palo Alto}
  \country{USA}
}
\email{stanfurd@stanford.edu}

\author{Ritvik Sharma}
\affiliation{
  \institution{Stanford University}
  \city{Stanford}
  \country{USA}
}
\email{rsharma3@stanford.edu}

\author{Marco Siracusa}
\affiliation{
  \institution{Barcelona Supercomputing Center}
  \city{Barcelona}
  \country{Spain}
}
\email{marco.siracusa@bsc.es}

\author{Fredrik Kjolstad}
\affiliation{
  \institution{Stanford University}
  \city{Stanford}
  \country{USA}
}
\email{kjolstad@stanford.edu}

\author{Kunle Olukotun}
\affiliation{
  \institution{Stanford University}
  \city{Stanford}
  \country{USA}
}
\email{kunle@stanford.edu}

\author{Olivia Hsu}
\affiliation{
  \institution{Stanford University}
  \city{Stanford}
  \country{USA}
}
\affiliation{
  \institution{Carnegie Mellon University}
  \city{Pittsburgh}
  \country{USA}
}
\email{owhsu@stanford.edu}
\renewcommand{\shortauthors}{Rubens Lacouture et al.}

\begin{abstract}
As deep learning models scale, sparse deep learning (DL) models that exploit sparsity in weights, activations, or inputs and specialized dataflow hardware have emerged as powerful solutions to address efficiency. We propose FuseFlow, a compiler that converts sparse machine learning models written in PyTorch to fused sparse dataflow graphs for reconfigurable dataflow architectures (RDAs). FuseFlow is the first compiler to support general cross-expression fusion of sparse operations. In addition to fusion across kernels (expressions), FuseFlow also supports optimizations like parallelization, dataflow ordering, and sparsity blocking. It targets a cycle-accurate dataflow simulator for microarchitectural analysis of fusion strategies. We use FuseFlow for design-space exploration across four real-world machine learning applications with sparsity, showing that full fusion (entire cross-expression fusion across all computation in an end-to-end model) is not always optimal for sparse models—fusion granularity depends on the model itself. FuseFlow also provides a heuristic to identify and prune suboptimal configurations. Using FuseFlow, we achieve performance improvements, including a ${\sim}$2.7x speedup over an unfused baseline for GPT-3 with BigBird block-sparse attention.
\end{abstract}




\maketitle % should come after the abstract

\section{Introduction}
\label{sec:intro}
\begin{figure}[t]
  \centering
  \revisionfig{\includegraphics[width=0.9\columnwidth]{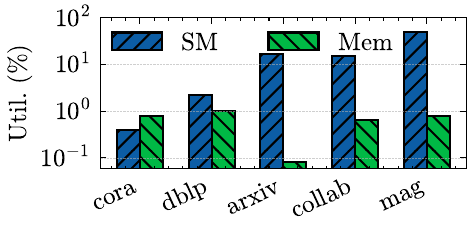}}
  \caption{\revision{Log plot of SM and DRAM utilization (\%) for \emph{PyG} GCN inference on an RTX~5090 across five datasets.}}
  
  \label{fig:gpu-util}
\end{figure}

% Sparse computation is a promising avenue for greater model efficiency in machine learning (ML) systems
% \revision{Sparse deep learning (DL)---models that exploit sparsity in weights, activations, or inputs---is a promising avenue for greater efficiency compared to dense deep learning} \cite{gale2020, gale2019state, han2015learningweightsconnectionsefficient, hoefler2021sparsity, wen2016learning}, often at the cost of lower hardware utilization. 
% \revision{Deep learning models may have sparse weights, activations, or inputs---naturally occurring or induced. Exploiting this sparsity during computation, which we call \emph{sparse deep learning (DL)}, offers greater efficiency than dense computation~\cite{gale2020, gale2019state, han2015learningweightsconnectionsefficient, hoefler2021sparsity, wen2016learning}.}
\revision{Deep learning models may have sparse weights, activations, or inputs---naturally occurring or induced. Exploiting this sparsity during computation, which we call \emph{sparse deep learning (DL)}, reduces compute and memory requirements but introduces irregular memory access patterns~\cite{gale2020, gale2019state, han2015learningweightsconnectionsefficient, hoefler2021sparsity, wen2016learning}.}
% \revision{Unlike dense deep learning, where all tensor elements are processed uniformly, sparse deep learning skips zero-valued computations, reducing both compute and memory requirements but introducing irregular memory access patterns.} 
To increase hardware efficiency, researchers are building specialized hardware to accelerate sparse computations~\cite{nowatzki2017, carsello2022amber, onyx, dadu2019spu, hegde2019extensor, fifer2021nguyen, triginst2013parashar, plasticine, rucker2021capstan,firoozshahian2023mtia,siracusa2023tensor}. In order to increase efficiency in these hardware architectures, they are also making increasing use of dataflow, or direct connections between coarse-grained functional units, to rely less on expensive caches, local memories, and memory operations. Dataflow architectures are particularly well-suited for sparse computation because they explicitly coordinate data movement through streaming connections rather than relying on caches, naturally handling the irregular memory access patterns inherent in sparse data~\cite{hsu2023sam,onyx,rucker2021capstan}. 
\revision{Empirically, GPUs are underutilized: A 3-layer GCN inference in \emph{PyTorch Geometric (PyG)} on an RTX~5090 across five real-world graphs shows consistently low compute (SM) utilization (avg $16.7\%$) and $\sim$1\% memory utilization (\Cref{fig:gpu-util}). These observations motivate specialized sparse dataflow accelerators and the compiler support to program them.}
% \revision{GPUs are poorly suited: GCN inference in \emph{PyG} on an RTX~5090 shows only 16.7\% average SM utilization and $\sim$1\% memory utilization across real-world graphs (\Cref{fig:gpu-util}), motivating specialized sparse dataflow accelerators and compiler support.}

\citet{hsu2023sam} introduced the Sparse Abstract Machine (SAM) to increase the programmability of these emerging sparse dataflow hardware architectures. The Sparse Abstract Machine is a dataflow abstract machine for sparse tensor algebra computations. It adopts a streaming dataflow model, where data flows between compute nodes. It can express any tensor algebra expression by composing a handful of simple and intuitive dataflow blocks. It also naturally supports expressing fused computation across multiple expressions, different ways to order the dataflow (the iteration order) within an expression (e.g., Gustavson's algorithm \cite{gustavson1978} versus inner product for sparse matrix multiplication), and lends itself to compile to fabricated hardware~\cite{onyx}. SAM is therefore a natural starting point as a compiler intermediate representation (IR) for targeting sparse ML models to dataflow hardware.

\citet{hsu2023sam} also describe a compilation flow from high-level sparse tensor algebra expressions to SAM dataflow graphs. Their Custard compiler generates SAM graphs that fuse operations across a sparse tensor algebra expression and lets users control the dataflow order. The Custard compilation algorithm is a significant step forward, as the first to demonstrate compilation of sparse tensor algebra expressions to dataflow. It is not, however, suitable for ML model compilation due to its limited capabilities for fusion. 

Because Custard compiles individual expressions, it is unable to fuse operations across expressions---a key feature in ML compilers~\cite{lacouturechallenges, Chen2018tvm}. Moreover, the intra-expression fusion that falls out of Custard's compilation algorithm fully fuses each expression without any support for partial fusion. Partial fusion is often desirable to provide some of the benefits of fusion while controlling the amount of reuse within a computation (as demonstrated by FlashAttention~\cite{dao2022flashattention}). \revision{The fusion--recomputation tradeoff is fundamental: fusion eliminates intermediate tensors between operations, reducing memory traffic, but excessive fusion can force recomputation of values~\cite{zhou2022react,dias2022sparselnr}. Conversely, insufficient fusion materializes intermediate tensors to memory, forcing more data movement.} This tradeoff is even more critical, and looks different from that of dense computation, since fused sparse computation may have better asymptotic complexity~\cite{bansal2023mosaic,hsu2023sam,kjolstad2020sparse,ahrens2022autoscheduling}.
Therefore, an ML compiler should expose the fusion granularity as a user schedule
so the fusion and reuse tradeoff can be explored across models. 

In this paper, we describe a new approach for lowering \revision{sparse DL models, models with one or more sparse tensors, to SAM dataflow graphs. Unlike prior work on sparse tensor algebra compilers that target individual expressions, \compiler{} compiles complete sparse DL inference pipelines, including nonlinear operations and masking.}
% general sparse ML workloads (beyond pure sparse tensor algebra) to SAM dataflow graphs. Sparse ML refers to deep learning models with one or more sparse tensors.
\revision{\compiler{} supports sparse tensors from any source, whether from pruning, natural zeros, or induced patterns, provided that the sparse data structure type is determined before compilation (\Cref{sec:sparsity-types}).}
% The supported sparse data structures include dense, COO, CSR, DCSR, n-D block structures, and combinations thereof for tensors of any dimensionality.

Our approach supports both cross-expression fusion and partial fusion, allowing users to explore the trade-off between fusion and reuse. 
Our work consists of two new IRs that enable this fusion exploration. The first IR, a fused Einstein Summation (Einsum) representation, tracks the flow of indexing (coordinate) data and values across fused expressions. Then, we introduce a new 
fusion table representation, a lowering IR that names and memoizes intermediate streams allowing the compiler to reference subgraphs before their materialization to efficiently emit the fused dataflow graph.

\revision{We also develop FuseFlow, the first academic end-to-end sparse ML compiler for reconfigurable dataflow architectures.}
FuseFlow compiles PyTorch~\cite{pytorch2} with sparse annotations~\cite{yan2024scorch,mpact} to SAM graphs. In FuseFlow, users leverage a scheduling language that lets them control fusion granularity and dataflow ordering of expressions. To support modern ML models, we also add support for dense blocks to support block-sparse tensors, non-linear functions, and masking operations to SAM.
Finally, our FuseFlow system can generate dataflow graphs that execute in a data-parallel fashion in addition to the pipeline parallelism native to dataflow graphs.
\revision{FuseFlow targets both cycle-accurate simulation and FPGA synthesis and existing dataflow accelerators~\cite{onyx, chen2025opal}, with validated agreement between FPGA and the simulation (\Cref{sec:hardware_validation}).}
Our technical contributions are thus:
\begin{itemize}
    \item \revision{A new data structure and} algorithm for fusion across multiple independent Einsum expressions (\Cref{sec:cross-expr-fusion}),
    \item \revision{A new abstraction that enables interleaved reductions for factored iteration and on-the-fly rearrangement of dataflow graphs (\Cref{sec:lowering}),}
    \item A lowering algorithm that converts the fused Einsum expressions to a dataflow representation (\Cref{sec:lowering})
    \item An end-to-end compiler framework for sparse dataflow machines (\Cref{sec:optimizations}). The implementation includes optimizations necessary for performant application code like parallelization, block sparsity, dataflow ordering, and a fusion heuristic.
\end{itemize}
We demonstrate the effectiveness of FuseFlow across four model classes by generating 56 equivalent dataflow configurations that yield speedups from \revision{${\sim}1.5$x} to ${\sim}3.9$x. Our evaluation underscores the \revision{importance} of selecting the \revision{appropriate} fusion granularity and shows that FuseFlow's heuristic successfully prunes inefficient configurations, offering critical insights for \revision{the deployment of} large-scale sparse ML applications on dataflow architectures.

\section{Sparse Abstract Machine Background} 
\label{sec:background}

\begin{figure}[t]
    \centering
        \includegraphics[width=0.95\columnwidth]{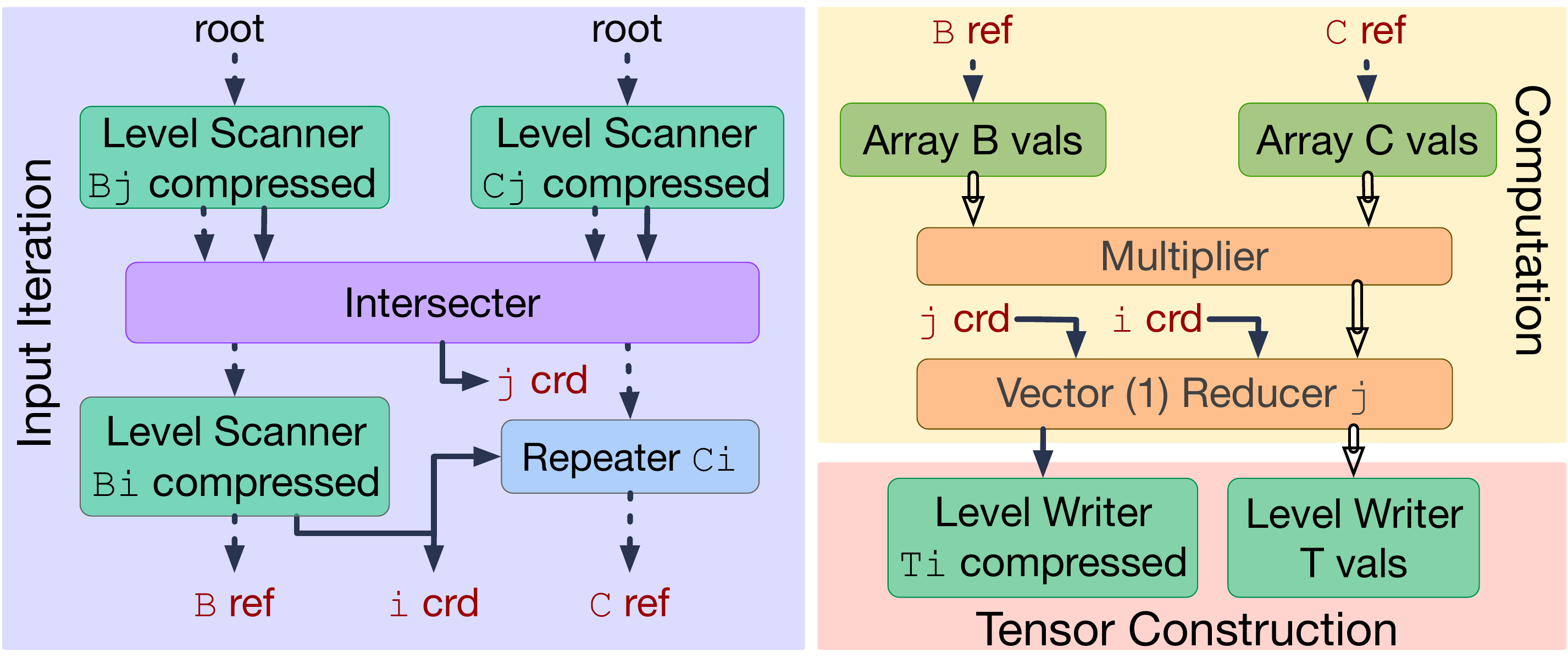}
        \caption{SAM graph for sparse-matrix vector multiplication with $j \rightarrow i$ dataflow.
        Streams: solid grey = coordinate (crd), dashed grey = reference (ref), double black = value (val).
        }
        \label{fig:sam-spmv}
    \label{fig:combined-spmv}
\end{figure}

We provide necessary background on the Sparse Abstract Machine (SAM)~\cite{hsu2023sam} to understand our \compiler{} system and the code that it generates. 
SAM expresses tensor algebra kernels as dataflow graphs by providing a streaming-tensor abstraction and primitives that compose to perform tensor algebra operations. Tensor algebra kernels can be expressed in Einsum notation where tensors are indexed by variables, with addition and multiplication as the core operations. The index variables specify how levels across tensors are broadcast, reduced, and contracted. SAM also introduces the Custard compiler, which compiles high-level Einsum into SAM dataflow graphs. These dataflow graphs are suitable for VLSI implementations and simulation but remain abstract in order to cleanly decouple programs from accelerator implementations.

\Cref{fig:sam-spmv} shows the SAM graph for sparse matrix-vector multiplication (SpMV) $T_i = B_{ij} C_j$ 
with the $j \rightarrow i$ dataflow ($\forall_{ji}~T_{i} \mathrel{+}= B_{ij}C_{j}$) where $B$ is stored in a compressed format (e.g., compressed sparse row (CSR))~\cite{chou2018,mpact}
SAM expresses tensors as streams of data with control tokens, where these tensor streams flow on the arrows between primitives (boxes) in a SAM dataflow graph. SAM’s primitives include:
\begin{description}[leftmargin=4pt]
    \item[Level scanners (LS)] traverse tensor levels. Nested LS produce streams that are logically equivalent to multidimensional tensors (e.g., $B_j$ with $B_i$ fetch matrix $B$’s coordinates). 
    \item[Stream joiners (Intersect/Union)] combine or skip coordinates across tensors (e.g., the $j$ intersect joins $B_j$ and $C_j$.
    \item[Repeaters (Rep)] broadcast operands (e.g., $C$ across each $i$).
    \item[ALU and reducers (Red)] \revision{perform elementwise operations} and reductions
    (e.g., reduce over $j$ in $j\!\rightarrow\! i$).
    \item[Level writers (LW) and coordinate droppers (CD)] write results and elide empty coordinates.
\end{description}
    
As in \Cref{fig:sam-spmv}, SAM primitives compose together with streams to form SAM graphs that represent any tensor algebra expression with varying dataflows.
Arrows in \Cref{fig:sam-spmv} connect dataflow primitives together and transmit streams, where each stream is a sequence of tokens that transmits one level of a tensor in fibertree form (a nested representation of per-level coordinates and values) \cite{sparseloop}.
Streams are of three types: coordinates (\texttt{crd}), references to inner levels (\texttt{ref}), and values (\texttt{val}). An $n$-order tensor is represented by $n$ coordinate streams plus one values stream ($n{+}1$ total).

SAM graphs comprise three regions (see shading in \Cref{fig:sam-spmv}): input iteration, computation, and tensor construction. The input iteration region (shown in blue) iterates through the tensor coordinates of all input operands, joining the sparse coordinates together (e.g. through intersecter$_j$). The computation region (shown in yellow) fetches data values using coordinates and computes the result values. And, the tensor construction region (shown in red) writes the result values and coordinates back to memory, dropping any zero coordinates. 
Our work builds upon SAM and addresses compiler limitations by targeting fused applications beyond single sparse tensor kernels.

\section{Forms of Fusion}
\label{sec:forms_fusion}

\begin{figure*}
    \centering
    \includegraphics[width=\linewidth]{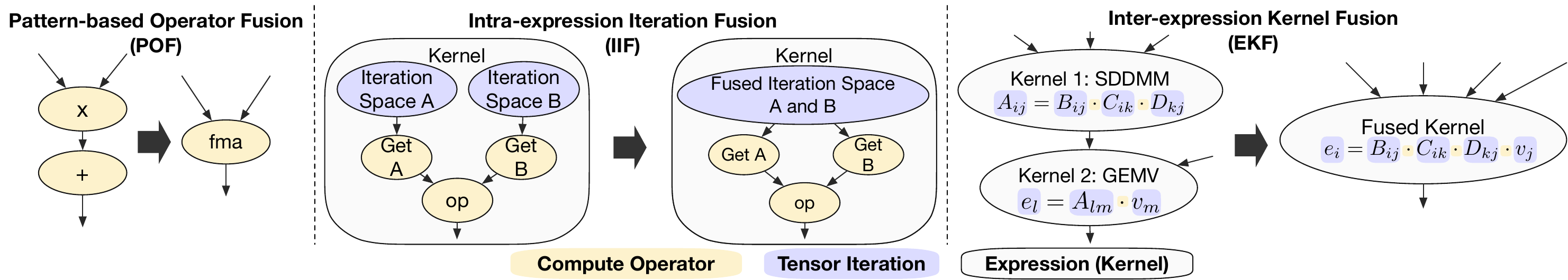}
    \caption{Dataflow diagrams for the forms of fusion, showing how they differ and are related.}
    \label{fig:fusion-forms}
\end{figure*}

Fusion in dense and sparse compilers can be categorized by scope and by technique. We describe three main types of fusion and provide a diagram of them in \Cref{fig:fusion-forms}. 

\begin{description}[leftmargin=8pt]
    \item[Pattern-based operator fusion (POF)] refers to merging operations based on recognized patterns, where sequences of operators are replaced by one kernel that fuses those operations. Often, the fused kernel is handcrafted. This operator fusion approach is common in dense compilers on CPUs, GPUs, and TPUs. Because POF is often completely automatic (all detected operator patterns are always replaced with the fused version), it is typically limited to localized patterns that are known in advance. Due to its localized nature, operator fusion is an intra-layer technique. 
    \item[Intra-expression iteration fusion (IIF)] 
    merges the iteration space of a single tensor algebra expression---co-iterating its inputs---without crossing kernel boundaries. IIF manifests as loop fusion for dense computation, co-iteration for sparse computation, and dataflow iteration fusion in dataflow graphs. Existing sparse tensor compilers that target dataflow hardware~\cite{hsu2023sam, hsu2025stardust} employ this type of fusion, co-iterating to generate fused iteration spaces that elide zeros for one sparse expression at a time.
    \item[Inter-expression kernel fusion (EKF)] fuses across different kernels or sub-computations into a single fused computation graph. This type of fusion can be implemented in conjunction with POF and IIF, but is not supported by existing sparse dataflow compilers~\cite{hsu2023sam,hsu2025stardust}.
\end{description}

Of the forms of fusion above, \Cref{tab:comparison} summarizes how prior frameworks fit into these categories. 
Existing frameworks thus leave a gap: no prior compiler automatically fuses across multiple sparse expressions in a general way. POF in dense compilers is limited to known templates and IIF in dense compilers is straightforward.  IIF in sparse compilers stop at single-kernel fusion. However, EKF is necessary to enable fusion across multiple sparse expressions in a model. FuseFlow addresses this gap by providing a general algorithm for fusing entire sparse ML pipelines across kernel boundaries. Therefore, as highlighted by \Cref{tab:comparison}, \compiler{} is the first sparse compiler to provide an algorithmic approach focusing on inter-expression kernel fusion. 

As \compiler{} is a sparse dataflow compiler, we contrast its fusion capabilities with the capabilities of the two prior sparse dataflow compilers Custard~\cite{hsu2023sam} and Stardust~\cite{hsu2025stardust} (C+S) in \Cref{fig:fuseflow_vs_cs}.\footnote{The motivation and description in \Cref{sec:intro} is also true for the Stardust compiler~\cite{hsu2025stardust}, another compiler from the same high-level sparse tensor algebra languages to a real dataflow accelerator~\cite{rucker2021capstan}.}
Custard and Stardust only support fusion within an expression and not across expressions. Although a user can combine expressions into a larger expression, which can then be fused, they must do so by hand and cannot fuse computations that have more than one result.
The various fused regions (blue boxes) compare C+S fusion regions with our \compiler{}, which fuses all kernels within a fusion region. 

% \revision{We quantitatively compare against C+S in \Cref{sec:priorcomparison}. \rl{move table back to this section}}
\compiler{}'s comprehensive fusion support leads to better performance.
In GCN on the OGB-Collab dataset~\cite{hu2020ogb} (\Cref{fig:fuseflow_vs_cs}b), fusion with Custard and Stardust using a handwritten rewrite yields 1.97$\times$ speedup over the unfused baseline. With less user effort, FuseFlow achieves another 1.33$\times$ speedup over C+S, leading to a $\sim$2.63$\times$ speedup in total.
\revision{We detail the comparison methodology for these results, and further analysis, in \Cref{sec:priorcomparison}.}
\compiler{}'s additional speedups come from its support for IIF during code generation. 
% Along with EKF, efficient fused sparse computation also hinges on the design of IIF, which we discuss next. 
Along with EKF, efficient fused sparse \revision{DL} also hinges on the design of IIF, which we discuss next. 

\begin{table}[]
  \centering
  \footnotesize
  \setlength{\tabcolsep}{3.0pt}
  \begin{tabular}{ccccc}
    \toprule
    \sffamily\bfseries{Tensor} & 
    \multirow{2}{*}{\shortstack[c]{\sffamily\bfseries{Multi-}\\\sffamily\bfseries{expression}}} &
    \multirow{2}{*}{\sffamily\bfseries{Sparsity}} &
    \sffamily\bfseries{Fusion} & 
    \multirow{2}{*}{\sffamily\bfseries{Backends}} \\
    \sffamily\bfseries{Compiler} & & & \sffamily\bfseries{Strategy} &  \\
    \midrule
    TensorRT~\cite{tensorrt} & \yes & \no & POF & GPU \\
    XLA~\cite{sabne2020xla} & \yes & \no & POF, IIF & CPU, GPU \\
    DNNFusion~\cite{niu2021dnnfusion} & \yes & \no & POF, IIF & CPU, GPU \\
    TVM~\cite{Chen2018tvm} & \yes & \no & POF, IIF & CPU, GPU, TPU \\
    TACO~\cite{kjolstad2017taco} & \no & \yes & IIF & CPU, GPU \\
    SparseTIR~\cite{sparsetir} & \no & \yes & IIF & CPU, GPU \\
    ReACT~\cite{zhou2022react} & \no & \yes & IIF & CPU, GPU \\
    Stardust~\cite{hsu2025stardust} & \no & \yes & IIF & Dataflow \\
    Custard~\cite{hsu2023sam} & \no & \yes & IIF & Dataflow \\
    \textbf{This Work} & \yes & \yes & EKF, IIF & Dataflow \\
    \bottomrule
  \end{tabular}
  \caption{{{Landscape of tensor compilers. \textbf{EKF (Inter-Expression Kernel Fusion)} enables fusion across multiple sparse tensor expressions. Prior sparse compilers only support \textbf{IIF (Intra-Expression Iteration Fusion)} within single kernels and dense compilers primarily rely on limited \textbf{POF (Pattern-based Operator Fusion)} via pattern matching.}}}
  \label{tab:comparison}
\end{table}

% \begin{figure}[t]
%   \centering
%     \includegraphics[width=0.85\columnwidth]{images/fuseflow-vs-cs.pdf}
%   \caption{Compiler fusion comparison with sparse attention. Custard+Stardust (C+S) support IIF and only support EKF manually where unsupported ops break EKF.}
%   \label{fig:fuseflow_vs_cs}
% \end{figure}

\begin{figure}[t]
  \centering
  \begin{subfigure}[c]{0.66\columnwidth}
    \centering
    \includegraphics[width=\linewidth]{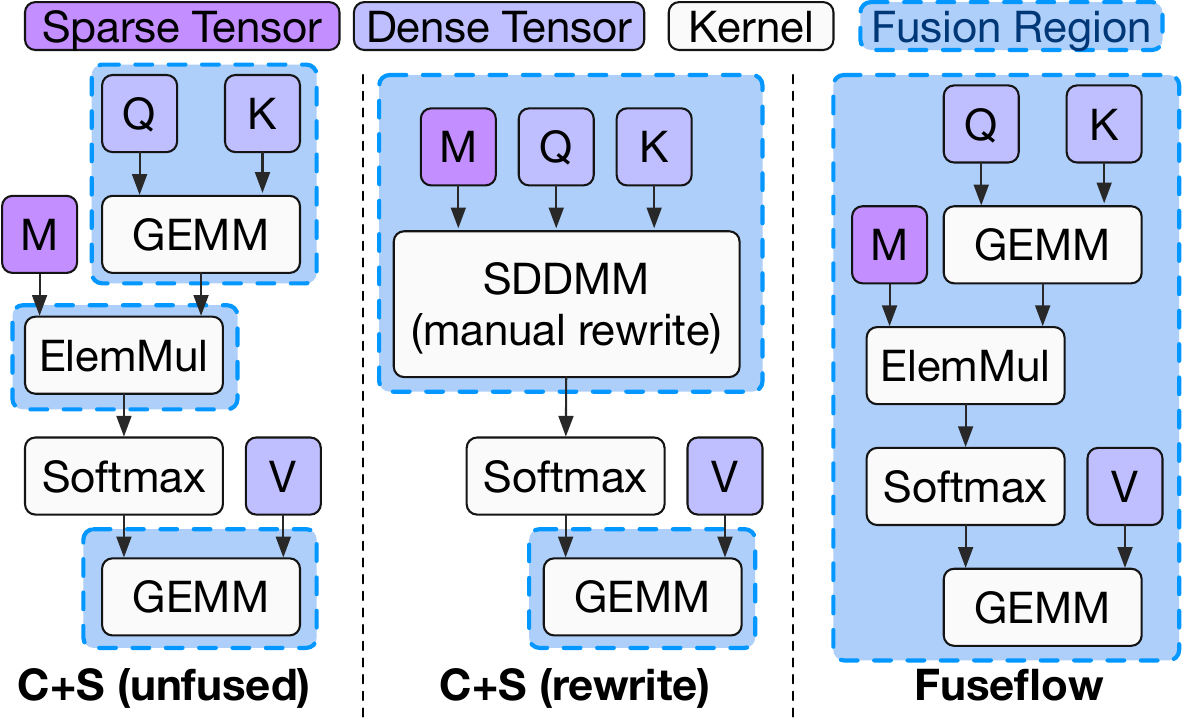}
    \caption{Compiler fusion comparison with sparse attention. Custard+Stardust (C+S) support IIF and only support EKF manually where unsupported ops break EKF.}
    \label{fig:fusion_coverage}
  \end{subfigure}%
  \hspace{0.6pt}
  \begin{subfigure}[c]{0.30\columnwidth}
    \centering
    \caption{Normalized GCN performance with compilers from (a).}
    \scriptsize
    \setlength{\tabcolsep}{1.0pt}
    \renewcommand{\arraystretch}{0.5}
    \begin{tabular}{@{}lc@{}}
      \toprule
      \multirow{2}{*}{\sffamily\bfseries Config}      & \sffamily\bfseries Speed- \\
          & \sffamily\bfseries up \\
      \midrule
      C+S (unfused)   & 1.00× \\
      C+S (rewrite) & 1.97× \\
      \textbf{FuseFlow} & \textbf{2.63×} \\
      \bottomrule
    \label{tab:comp_speedup}
    \end{tabular}
    \end{subfigure}
  \caption{Comparing fusion coverage and performance.}
  \label{fig:fuseflow_vs_cs}
\end{figure}

\paragraph*{The Iteration Space Problem}
\label{sec:iter_problem}
{In both inter- and intra-expression fusion approaches for sparse \revision{tensor operations}, there are multiple equivalent ways to iterate through sparse tensors, each with fundamental tradeoffs. Two primary costs, coordinate processing and computation, define the tradeoff between globally fused and \factored{} iteration spaces.} 

{A globally fused iteration space, shown in \Cref{fig:fullyfused}, iterates over every index variable, creating an n-dimensional iteration space, where n is the number of index variables {(e.g., 4-dimensional in \Cref{fig:fullyfused})}. It efficiently filters unnecessary numerical computations but incurs significant coordinate processing overhead, causing coordinate explosion as expressions grow. Prior work on sparse tensor algebra compilation to dataflow accelerators by default generates code that traverses a global iteration space~\cite{hsu2023sam,hsu2025stardust}.}
In contrast, \factored{} iteration iterates pairwise over input tensors (see \Cref{fig:binaryiterationspace}). It generates multiple smaller sub-spaces, one per binary operation (i.e. two 3-dimensional iteration spaces in \Cref{fig:binaryiterationspace}).
Each sub-space independently handles coordinate processing, significantly reducing overhead by limiting coordinate analysis to binary operations. However, as this analysis is local rather than global, \factored{} iteration may miss opportunities to skip unnecessary computations, potentially increasing operations.

\begin{figure}[]
    \centering
    % \scriptsize
  \begin{subfigure}[b]{.4\textwidth}
    \centering
    % \begin{lstlisting}
    \setminted{escapeinside=||}
    \begin{minted}[fontsize=\small, linenos, xleftmargin=0pt]{c}
    for(int i = 0; i < I; i++)
      |\highlight{for(int k = 0; k < K; k++)}|
        |\highlight{for(int j = 0; j < J; j++)}|
          for(int l = 0; l < L; l++)
            D[i,l] += A[i,k] * B[k,j] * C[j,l]
    \end{minted} 
    % \end{lstlisting} 
    \caption{Global iteration space with loops }
    \label{fig:fullyfused}
  \end{subfigure}
  \hfill
  \begin{subfigure}[b]{.4\textwidth}
    % \begin{lstlisting}
    \setminted{escapeinside=||}
    \begin{minted}[fontsize=\small, linenos, xleftmargin=0pt]{c}
    for(int i = 0; i < I; i++)
      |\highlight{for(int k = 0; k < K; k++)}|
        for(int j = 0; j < J; j++)
          E[i,j] += A[i,k] * B[k,j]
      |\highlight{for(int j = 0; j < J; j++)}|
        for(int l = 0; l < L; l++)
          D[i,l] += E[i,j] * C[j,l]
    \end{minted} 
    % \end{lstlisting} 
    \caption{\Factored{} iteration space with loops}
    \label{fig:binaryiterationspace}
  \end{subfigure}
  \caption{Two iteration patterns for {$\forall_{ikjl} C_{il} \mathrel{+}= A_{ik}B_{kj}C_{jl}$}, that are represented via loop nests with higher-order reduction variables highlighted in blue~\cite{kjolstad2019workspaces}. \compiler{} lowers to a dataflow input iteration graph with a \factored{} iteration space (b), whereas prior work produces dataflow graphs with fully fused iteration spaces (a). }
  \label{fig:loopnests}
  %%\vspace{-2.0em}
\end{figure}
% \begin{figure}[]
%     \centering
%     \scriptsize
%   \begin{subfigure}[b]{.30\textwidth}
%     \centering
% \begin{lstlisting}[basicstyle=\ttfamily\footnotesize,numbers=left,xleftmargin=0pt,escapeinside={(*@}{@*)}]
% for(int i = 0; i < I; i++)
%   (*@\colorbox{blue!20}{for(int k = 0; k < K; k++)}@*)
%     (*@\colorbox{blue!20}{for(int j = 0; j < J; j++)}@*)
%       for(int l = 0; l < L; l++)
%         D[i,l] += A[i,k] * B[k,j] * C[j,l]
% \end{lstlisting}
%     \caption{Global iteration space with loops }
%     \label{fig:fullyfused}
%   \end{subfigure}
%   \begin{subfigure}[b]{.30\textwidth}
% \begin{lstlisting}[basicstyle=\ttfamily\footnotesize,numbers=left,xleftmargin=0pt,escapeinside={(*@}{@*)}]
% for(int i = 0; i < I; i++)
%   (*@\colorbox{blue!20}{for(int k = 0; k < K; k++)}@*)
%     for(int j = 0; j < J; j++)
%       E[i,j] += A[i,k] * B[k,j]
%   (*@\colorbox{blue!20}{for(int j = 0; j < J; j++)}@*)
%     for(int l = 0; l < L; l++)
%       D[i,l] += E[i,j] * C[j,l]
% \end{lstlisting}
%     \caption{\Factored{} iteration space with loops}
%     \label{fig:binaryiterationspace}
%   \end{subfigure}
%   \caption{Two iteration patterns for {$\forall_{ikjl} C_{il} \mathrel{+}= A_{ik}B_{kj}C_{jl}$}, that are represented via loop nests with higher-order reduction variables highlighted in blue~\cite{kjolstad2019workspaces}. \compiler{} lowers to a dataflow input iteration graph with a \factored{} iteration space (b), whereas prior work produces dataflow graphs with fully fused iteration spaces (a). }
%   \label{fig:loopnests}
% \end{figure}

{Global iteration spaces often perform poorly for sparse ML applications for two key reasons. First, these applications typically contain numerous higher-order tensors and indices, leading to a dimensionality explosion. Second, the mixture of sparse and dense tensors increases iteration points within each dimension. This combination makes traversing sparse ML models with global iteration significantly less efficient than \factored{} iteration approaches.
Hence, we opt for \factored{} input-iteration in \compiler{}, by design, when pushing fusion through our lowering algorithm.}

% Most/all of these Macros are taken from Fred's thesis

% Colors
\definecolor{overviewgray}  {RGB}{219,219,235}
\definecolor{thesisblue}  {RGB}{0,80,235}
\definecolor{thesispurple}{RGB}{176,095,183}
\definecolor{thesisorange}{RGB}{252,128,8}
\definecolor{thesisgreen} {RGB}{0,235,80}
\definecolor{thesisred}   {RGB}{231,01,03}
\definecolor{thesismaroon}   {RGB}{175,012,035}
\definecolor{thesisgray}  {RGB}{234,234,241}
\definecolor{thesisdarkgray}  {RGB}{80,80,80}
\definecolor{aqua}{RGB}{0,255,255}
\newcommand\thesisblue[1]{\textcolor{thesisblue}{#1}}
\newcommand\thesispurple[1]{\textcolor{thesispurple}{#1}}
\newcommand\thesisorange[1]{\textcolor{thesisorange}{#1}}
\newcommand\thesisgreen[1]{\textcolor{thesisgreen}{#1}}
\newcommand\thesisred[1]{\textcolor{thesisred}{#1}}
\newcommand\thesisgray[1]{\textcolor{thesisgray}{#1}}
\newcommand\thesisdarkgray[1]{\textcolor{thesisdarkgray}{#1}}

% Iteration graphs
\def\modeuniverse{$\mathbb{U}$}
\colorlet{igpathA}{thesisblue}
\colorlet{igpathB}{thesisred}
\colorlet{igpathC}{thesisgreen}
\colorlet{igpathD}{thesisorange}
\colorlet{igdim}  {black}
\newcommand\igpathA[1]{\textcolor{igpathA}{#1}}
\newcommand\igpathB[1]{\textcolor{igpathB}{#1}}
\newcommand\igpathC[1]{\textcolor{igpathC}{#1}}
\newcommand\igpathD[1]{\textcolor{igpathD}{#1}}
\newcommand\igdim[1]  {\textcolor{igdim}{\modeuniverse_{#1}}}
\tikzstyle{ignode}  = [circle, draw=black, thin, minimum size=6.5mm]
\tikzstyle{igedge}  = [-latex, line width=1pt]
\tikzstyle{igop}    = []
\tikzstyle{igpathA} = [color=igpathA]
\tikzstyle{igpathB} = [color=igpathB]
\tikzstyle{igpathC} = [color=igpathC]
\tikzstyle{igpathD} = [color=igpathD]
\colorlet{igpathAB}{thesispurple}

% Venn diagrams
\colorlet{vennAB}{igpathAB!50!white}
\colorlet{vennA}{igpathA!50!white}
\colorlet{vennB}{igpathB!50!white}
\colorlet{vennC}{igpathC!50!white}
\colorlet{vennD}{igpathD!50!white}
\tikzstyle{venncircle} = [line width=1pt]
\def\venn{circle (1.2)}
\def\binaryvennA{(180:0.6) \venn}
\def\binaryvennB{(000:0.6) \venn}
\def\tertiaryvennC{(150:0.61) \venn}
\def\tertiaryvennD{(030:0.61) \venn}
\def\tertiaryvennE{(270:0.61) \venn}
\newcommand\VennScale{0.26}

\def\smallvenn{circle (1.1)}
\def\smallbinaryvennA{($ (0,1) + (180:0.5) $) \smallvenn}
\def\smallbinaryvennB{($ (0,1) + (000:0.5) $) circle (1.1) \smallvenn}
\def\smalltertiaryvennC{($(0,1) + (150:0.61) $) \smallvenn}
\def\smalltertiaryvennD{($(0,1) + (030:0.61) $) \smallvenn}
\def\smalltertiaryvennE{($(0,1) + (270:0.61) $) \smallvenn}
\def\spacegridscalesmall{1.2}
\def\spacegridscalexs{0.7}
\def\spacegridscale{1.5}
\colorlet{pointcolor}{black!80!white}
\colorlet{coordcolor}{black!90!white}
\tikzstyle{spacegrid} = [gray, thick]
\def\spacepointsize{\tiny}
\NewDocumentCommand{\spacepoint}{ O{0,0} O{} m m m }{
  \filldraw [pointcolor,#2] ($ (#1) + (#4, #3-#5) $) circle [radius=3pt];
  \node at ($ (#1) + (#4, #3-#5) + (0.4,0.25)$) {{\spacepointsize \textcolor{coordcolor}{$(#5,#4)$}}};
}
\newcommand\included[1]{{\spacepointsize #1}}
\newcommand\excluded[1]{{\spacepointsize \thesisdarkgray{#1}}}

% Lattices
\tikzset{cross/.style={cross out, draw=black, minimum size=2*(#1-\pgflinewidth), inner sep=0pt, outer sep=0pt},
%default radius will be 1pt. 
cross/.default={1pt}}

\colorlet{highlightcolor}{thesisgray}
\definecolor{darkhighlightcolor}{gray}{0.80}
\tikzstyle{highlightstyle} = [draw=darkhighlightcolor,fill=highlightcolor,rounded corners=5pt,line width=0.1pt]

\def\latticedim {\modeuniverse}
\def\lplevelone{0}
\def\lpleveltwo{-1}
\def\lplevelthree{-2.0}
\def\lplevelfour{-3}
\def\lplevelfive{-4}
\def\lptext{0.8}
\tikzstyle{lpedge} = []%line width=1pt]
\def\lpheight{6}
\def\lplinegap{0.8}
\tikzstyle{lpstyle} = [
  draw,
  fill=thesisgray,
  rounded rectangle,
  minimum width=1.5cm,
  minimum height = \lpheight{}mm,
  inner xsep=5,
  outer sep=0,
]

\tikzstyle{lpstylenew} = [
  draw,
  fill=#1,
  text opacity = 1.0,
  rounded rectangle,
  fill opacity = 0.6,
  minimum width=1.5cm,
  minimum height = \lpheight{}mm,
  inner xsep=6,
  outer sep=0,
]

\tikzstyle{highlightlpstyle} = [
  draw=thesisgreen!50!black,
  fill=#1,
  rounded rectangle,
  minimum width=1.5cm,
  minimum height = \lpheight{}mm,
  inner xsep=5,
  outer sep=0,
]

\tikzstyle{bluelpstyle} = [
  draw=blue,
  fill=thesisgray,
  rounded rectangle,
  minimum width=2.0cm,
  minimum height = \lpheight{}mm,
  inner xsep=5,
  outer sep=0,
]

\tikzstyle{lpestyle} = [near start]
\def\lpbar{\; \mid \;}
\newcommand{\latticepoint}[4]{
  \def\temp{#4}\ifx\temp\empty
    \node[lpstyle] (#1) at #2 {#3};
  \else
    \node[lpstyle] (#1) at #2 {#3 $\; \mid \;$ #4};
  \fi
}

\newcommand{\latticepointcolor}[5]{
  \def\temp{#4}\ifx\temp\empty
    \node[lpstylenew={#5}] (#1) at #2 {#3};
  \else
    \node[lpstylenew={#5}] (#1) at #2 {#3 $\; \mid \;$ #4};
  \fi
}

\newcommand{\highlightedlatticepoint}[5]{
  \def\temp{#4}\ifx\temp\empty
    \node[highlightlpstyle={#5}, ultra thick] (#1) at #2 {#3};
  \else
    \node[highlightlpstyle={#5}, ultra thick] (#1) at #2 {#3 $\; \mid \;$ #4};
  \fi
}

\newcommand{\bluelp}[4]{
  \def\temp{#4}\ifx\temp\empty
    \node[bluelpstyle] (#1) at #2 {#3};
  \else
    \node[bluelpstyle] (#1) at #2 {#3 $\; \mid \;$ #4};
  \fi
}

\tikzset{
    actor/.style={
      draw,
      rounded rectangle,
      minimum width=1.5cm,
      minimum height = \lpheight{}mm,
      text opacity = 1,
      inner xsep=5,
      outer sep=0
    },
    actor crossed out/.style={
        actor, 
        append after command={
            node [
                opacity=0.35,
                fit=(\tikzlastnode),
                draw=red,
                ultra thick,
                inner sep=-5,
                cross out
            ] {}
        }
    }
}

\tikzset{
    actor2/.style={
      draw=orange,
      rounded rectangle,
      minimum width=2.0cm,
      minimum height = \lpheight{}mm,
      text opacity = 1,
      inner xsep=5,
      outer sep=0
    },
    actor2 crossed out/.style={
        actor2, 
        append after command={
            node [
                opacity=0.3,
                fit=(\tikzlastnode),
                draw=red,
                thick,
                inner sep=-5,
                cross out
            ] {}
        }
    }
}

\tikzset{
    actor3/.style={
      draw=blue,
      rounded rectangle,
      minimum width=2.0cm,
      minimum height = \lpheight{}mm,
      text opacity = 1,
      inner xsep=5,
      outer sep=0
    },
    actor3 crossed out/.style={
        actor3, 
        append after command={
            node [
                opacity=0.3,
                fit=(\tikzlastnode),
                draw=red,
                thick,
                inner sep=-5,
                cross out
            ] {}
        }
    }
}

\newcommand{\omitpoint}[4]{
  \def\temp{#4}\ifx\temp\empty
    \node[actor crossed out] (#1) at #2 {#3};
  \else
    \node[actor crossed out] (#1) at #2 {#3 $\; \mid \;$ #4};
  \fi
}

\newcommand{\homitpoint}[4]{
  \def\temp{#4}\ifx\temp\empty
    \node[actor2 crossed out] (#1) at #2 {#3};
  \else
    \node[actor2 crossed out] (#1) at #2 {#3 $\; \mid \;$ #4};
  \fi
}

\newcommand{\bhomitpoint}[4]{
  \def\temp{#4}\ifx\temp\empty
    \node[actor3 crossed out] (#1) at #2 {#3};
  \else
    \node[actor3 crossed out] (#1) at #2 {#3 $\; \mid \;$ #4};
  \fi
}

\newcommand{\latticebottom}[1]{
  \node[draw, lpstyle] at #1 (bottom) {$\emptyset$};
}

\newcommand{\latticeinline}[1]{\mbox{$(\hspace{0.2mm}$#1$\hspace{0.2mm})$}}
\newcommand{\lpinline}[1]{\,\tikz[overlay]\node[draw=black!50, inner sep=1.6pt,
anchor=text, rectangle, rounded corners=1mm, line width=0.22mm]
{#1};\phantom{#1}\,}
\newcommand{\scalelattice}{0.14}

% Comparision of grid of points

\newcommand\figIterationSpacesDense {
    \begin{tikzpicture}[scale=\spacegridscalexs]
      \draw [spacegrid] (0,0) grid (3,2);
      \foreach \x in {0,...,3}
        \foreach \y in {0,...,2} {
          \spacepoint{2}{\x}{\y}
        }
    \node[black,  label=$A_{ik}$]  at (1.5,-1) {};
    \end{tikzpicture}
    % \subcaption{A_{ik}}
}

\newcommand\figIterationSpacesSparse {
    \begin{tikzpicture}[scale=\spacegridscalexs]
      \draw [spacegrid] (0,0) grid (3,2);
      \spacepoint{2}{0}{0}
      \spacepoint{2}{1}{0}
      \spacepoint{2}{1}{2}
      \spacepoint{2}{2}{1}
      \spacepoint{2}{3}{2}
    \node[black,  label=$B_{kj}$]  at (1.6,-1) {};
    \end{tikzpicture}
    % \subcaption{B_{kj}}
}

\newcommand\figIterationSpacesRandSparse {
    \begin{tikzpicture}[scale=\spacegridscalexs]
      \draw [spacegrid] (0,0) grid (3,2);
      \spacepoint{2}{0}{0}
      \spacepoint{2}{1}{2}
      \spacepoint{2}{1}{1}
      \spacepoint{2}{2}{2}
      \spacepoint{2}{3}{2}
    \node[black,  label=$C_{jl}$]  at (1.6,-1) {};
    \end{tikzpicture}
    % \subcaption{A_{jl}}
}

% Start the TikZ figure

\newcommand\figStackedCubes {
% Two cubes
\begin{tikzpicture}[node distance=2cm, scale=.45]

% Example nodes and lines
\coordinate (a0) at (0,0);
\coordinate (a1) at (0,3);
\coordinate (a2) at (3,0);
\coordinate (a3) at (3,3);

\coordinate (b0) at (1,1);
\coordinate (b1) at (1,4);
\coordinate (b2) at (4,1);
\coordinate (b3) at (4,4);

% define offset for second set of nodes
% define variable for offset
\pgfmathsetmacro{\offsetx}{ 3.4}
\pgfmathsetmacro{\offsety}{-2.0}

\coordinate (c0) at (0 + \offsetx,0 + \offsety);
\coordinate (c1) at (0 + \offsetx,3 + \offsety);
\coordinate (c2) at (3 + \offsetx,0 + \offsety);
\coordinate (c3) at (3 + \offsetx,3 + \offsety);

\coordinate (d0) at (1 + \offsetx,1 + \offsety);
\coordinate (d1) at (1 + \offsetx,4 + \offsety);
\coordinate (d2) at (4 + \offsetx,1 + \offsety);
\coordinate (d3) at (4 + \offsetx,4 + \offsety);

% Lines for each 3D cube 
\foreach {\i} in {a, b, c, d} {
    \draw [spacegrid] (\i0) -- (\i1);
    \draw [spacegrid] (\i1) -- (\i3);
    \draw [spacegrid] (\i3) -- (\i2);
    \draw [spacegrid] (\i2) -- (\i0);
}

\foreach {\i} in {0, 1, 2, 3} {
    \draw [spacegrid] (a\i) -- (b\i);
    \draw [spacegrid] (c\i) -- (d\i);
}

% Diagonal dotted lines
\foreach {\i} in {0, 1, 2, 3} {
    \draw[dashed, spacegrid] (a\i) -- (c\i);
    \draw[dashed, spacegrid] (b\i) -- (d\i);
}
% Dots at each coordinate
\foreach {\i} in {0,1,2,3} {
    \fill (a\i) circle (4pt);
    \fill (b\i) circle (4pt);
    \fill (c\i) circle (4pt);
    \fill (d\i) circle (4pt);
}

\node[black,  label=Global Iteration Space]  at (4.0,-4.5) {};
\end{tikzpicture}
}

\newcommand\figStackedOverlapCubes {
% Two cubes
\begin{tikzpicture}[node distance=2cm, scale=.45]

% Example nodes and lines
\coordinate (a0) at (0,0);
\coordinate (a1) at (0,3);
\coordinate (a2) at (3,0);
\coordinate (a3) at (3,3);

\coordinate (b0) at (1,1);
\coordinate (b1) at (1,4);
\coordinate (b2) at (4,1);
\coordinate (b3) at (4,4);

% define offset for second set of nodes
% define variable for offset
\pgfmathsetmacro{\offsetx}{ 0.0}
\pgfmathsetmacro{\offsety}{-3.0}

\coordinate (c0) at (0 + \offsetx,0 + \offsety);
\coordinate (c1) at (0 + \offsetx,3 + \offsety);
\coordinate (c2) at (3 + \offsetx,0 + \offsety);
\coordinate (c3) at (3 + \offsetx,3 + \offsety);

\coordinate (d0) at (1 + \offsetx,1 + \offsety);
\coordinate (d1) at (1 + \offsetx,4 + \offsety);
\coordinate (d2) at (4 + \offsetx,1 + \offsety);
\coordinate (d3) at (4 + \offsetx,4 + \offsety);

% Lines for each 3D cube 
\foreach {\i} in {a, b, c, d} {
    \draw [spacegrid] (\i0) -- (\i1);
    \draw [spacegrid] (\i1) -- (\i3);
    \draw [spacegrid] (\i3) -- (\i2);
    \draw [spacegrid] (\i2) -- (\i0);
}

\foreach {\i} in {0, 1, 2, 3} {
    \draw [spacegrid] (a\i) -- (b\i);
    \draw [spacegrid] (c\i) -- (d\i);
}

% Dots at each coordinate
\foreach {\i} in {0,1,2,3} {
    \fill (a\i) circle (4pt);
    \fill (b\i) circle (4pt);
    \fill (c\i) circle (4pt);
    \fill (d\i) circle (4pt);
}

% Diagonal dotted lines
% \foreach {\i} in {0, 1, 2, 3} {
    % \draw[dashed] (a\i) -- (c\i);
    % \draw[dashed] (b\i) -- (d\i);
% }
\node[black,  label=Binary Iteration Space]  at (1.6,-4.5) {};

\end{tikzpicture}
}

\newcommand\figIterationSpaceComparison {
\begin{figure*}
    \centering
    \begin{minipage}{0.15\linewidth}
        \centering
    	\figIterationSpacesDense
	\end{minipage}
	\hfill
    \begin{minipage}{0.15\linewidth}
        \centering
    	\figIterationSpacesSparse
	\end{minipage}
	\hfill
    \begin{minipage}{0.15\linewidth}
        \centering
    	\figIterationSpacesRandSparse
	\end{minipage}
    \hfill
    \begin{minipage}{0.15\linewidth}
        \centering
    	\figStackedCubes
	\end{minipage}
    \hfill
    \begin{minipage}{0.15\linewidth}
        \centering
    	\figStackedOverlapCubes
	\end{minipage}
    \caption{
        A grid representation of iteration spaces showing a dense and sparse iteration space for $4 \times 3$ matrix. \todo{Might need to change iteration spaces to 2x2, and figure out where points should be}
        \label{fig:grid_iter_spaces}
    }
\end{figure*}
}

\newcommand\figIterationSpaceSet {
\begin{figure}
    \centering
    \figCoordVenn
    \caption{\TODO {Maybe merge with point set picture}. Venn Diagram showing which sets tensor coordinates belong to. All the coordinates belonging to A are in the subset A within the universe of coordinates. This is a Venn Diagram representation of \figref{iteration-spaces-sparse}}.
    \label{fig:venn_iter_space}
\end{figure}
}

\newcommand\figIterSpaceExamples{
\begin{figure}
\begin{minipage}[c]{0.5\textwidth}
\centering
\small
    \begin{tikzpicture}[scale=1.25]
    \footnotesize
    \draw[] (-2.25, -2.25) rectangle (2.25, 1.75) node [text=black,below left] {\large\modeuniverse};
    \fill[white] \tertiaryvennD;
    \begin{scope}[blend group=overlay]
        \clip \tertiaryvennD;
        \fill[vennA, fill opacity=0.7] \tertiaryvennE;
        \fill[thesispurple, fill opacity=0.7] \tertiaryvennC;
    \end{scope}
    \fill[white] \tertiaryvennE;
    \begin{scope}[blend group=overlay]
        \clip \tertiaryvennC;
        \clip \tertiaryvennD;
        \clip \tertiaryvennE;
        \fill[thesispurple, fill opacity=1] \tertiaryvennC;
        \fill[vennC, fill opacity=0.5] \tertiaryvennC;
    \end{scope}
    \draw[venncircle, draw=vennA] \tertiaryvennC;
    \draw[venncircle, draw=vennB] \tertiaryvennD;
    \draw[venncircle, draw=vennC] \tertiaryvennE;
    \node[rotate=60] at (150:1.20) {$\text{max}(A, \infty, 0)$};
    \node[rotate=-60] at (030:1.20) {$\text{max}(\infty, B, 0)$};
    \node at (270:1.20) {$\text{max}(\infty, \infty, C)$};
    \node at (090:0.95) {max};
    \node at (090:0.75) {$(A, B, 0)$};
    \node at (210:0.9) {max};
    \node at (220:0.95) {$(A, \infty, C)$};
    \node at (330:0.9) {\text{max}};
    \node at (320:0.95) {$(\infty, B, C)$};
    \node at (090:0.00) {$\text{max}(A, B, C)$};
    \node at (310:2.5) {$\max(\infty, \infty, 0)$};
    \node[text=thesisgreen, ultra thick] at (270:2) {\large$C$};
    \node[text=thesisblue, ultra thick] at (150:2) {\large$A$};
    \node[text=thesisred, ultra thick] at (030:2) {\large$B$};
\end{tikzpicture}
\caption{Illustration of case (1), where $f$ is the ternary max operator, A and B have fill value $\infty$ and C has fill value 0.}
\label{fig:annihilator-example}
\end{minipage}%
\begin{minipage}[c]{0.5\textwidth}
\begin{minipage}[c]{\textwidth}
\centering
\small
\begin{tikzpicture}[scale=1]
  \begin{scope}[blend mode = overlay]
    \clip \smallbinaryvennA;
    \fill[vennA] \smallbinaryvennA;
  \end{scope}
  \begin{scope}[blend mode = overlay]
    \clip \smallbinaryvennB;
    \fill[vennB] \smallbinaryvennB;
  \end{scope}
  \begin{scope}[blend mode = overlay]
    \clip \smallbinaryvennA;
    \fill[thesispurple] \smallbinaryvennB;
  \end{scope}
  \draw[venncircle, draw=vennA] \smallbinaryvennA;
  \draw[venncircle, draw=vennB] \smallbinaryvennB;
  \draw (-2, -0.75) rectangle (2.25, 2.5) node [text=black,below left] {\modeuniverse};
  \node at ($ (0,1) + ( 170:1.1)$) {min};
  \node at ($ (0,0.8) + (180:1.15)$) {$(A, v)$};
  \node at ($ (0,1.2) $) {min};
  \node at ($ (0,0.8) $) {$(A, B)$};
  \node at ($ (0,1) + ( 10:1.1)$) {min};
  \node at ($ (0,0.8) + (  0:1.1)$) {$(v,B)$};
  \node at ($ (0,1) + ( 330:2.05)$) {min};
  \node at ($ (0,1) + ( 320:2.25)$) {$(v, v)$};
  \node[text=thesisblue] at ($ (0,1) + (  240:1.5)$) {\large$A$};
  \node[text=thesisred] at ($ (0,1) + (  300:1.5)$) {\large$B$};
\end{tikzpicture}
\caption{Illustration of case (2), where $f$ is the idempotent min operator and all arguments have the same fill value $v$.}
\label{fig:idempotent-example}
\end{minipage}
\begin{minipage}[c]{\textwidth}
\vspace{0.5em}
\centering
\small
\begin{tikzpicture}[scale=1]
  \begin{scope}[blend mode = overlay]
    \clip \smallbinaryvennA;
    \fill[vennA] \smallbinaryvennA;
  \end{scope}
  \begin{scope}[blend mode = overlay]
    \clip \smallbinaryvennB;
    \fill[vennB] \smallbinaryvennB;
  \end{scope}
  \begin{scope}[blend mode = overlay]
    \clip \smallbinaryvennA;
    \fill[thesispurple] \smallbinaryvennB;
  \end{scope}
  \draw[venncircle, draw=vennA] \smallbinaryvennA;
  \draw[venncircle, draw=vennB] \smallbinaryvennB;
  \draw (-2, -0.75) rectangle (2.25, 2.5) node [text=black,below left] {\modeuniverse};
  \node at ($ (0,1) + ( 170:1.1)$) {max};
  \node at ($ (0,0.8) + (180:1.15)$) {$(A, 42)$};
  \node at ($ (0,1.2) $) {max};
  \node at ($ (0,0.8) $) {$(A, B)$};
  \node at ($ (0,1) + ( 10:1.1)$) {max};
  \node at ($ (0,0.8) + (  0:1.1)$) {$(-\infty,B)$};
  \node at ($ (0,1) + ( 330:2)$) {max};
  \node at ($ (0,1) + ( 320:2.2)$) {$(-\infty, 42)$};
  \node[text=thesisblue] at ($ (0,1) + (240:1.5)$) {\large$A$};
  \node[text=thesisred] at ($ (0,1) + (300:1.5)$) {\large$B$};
\end{tikzpicture}
\caption{Illustration of case (3), where $f$ is the max operator with identity $-\infty$, A has fill value 42 and B has fill value $-\infty$.}
\label{fig:identity-example}
\end{minipage}
\end{minipage}
\end{figure}
}

Avoiding global iteration space materialization requires a complete restructuring of the dataflow graph and its compiler lowering. \Factored{} iteration preserves the order of reduction operations, unlike global iteration. \Cref{fig:loopnests} highlights these behaviors. Recall that SAM graphs comprise three sequential regions—input iteration, computation, and tensor construction (see \Cref{sec:background}). Global iteration computations occur at the innermost loop (line 5 in \Cref{fig:fullyfused}), while \factored{} iteration interleaves loops and computations (lines 4 and 7 in \Cref{fig:binaryiterationspace}). From a dataflow perspective, rather than loop transformations, the graph input iteration and computation pipelines need to be interleaved, which we show later in \Cref{fig:lowering-comparison}.
Specifically, higher-order reducers produce coordinate streams that must interact with the input iterations of subsequent operations to enable fusion. Therefore, we need a new compiler approach, like \compiler{}'s, to handle efficient sparse ML workloads on dataflow hardware. 

\section{Overview of FuseFlow}
\label{sec:overview}

\begin{figure*}[]
\includegraphics[width=\linewidth]{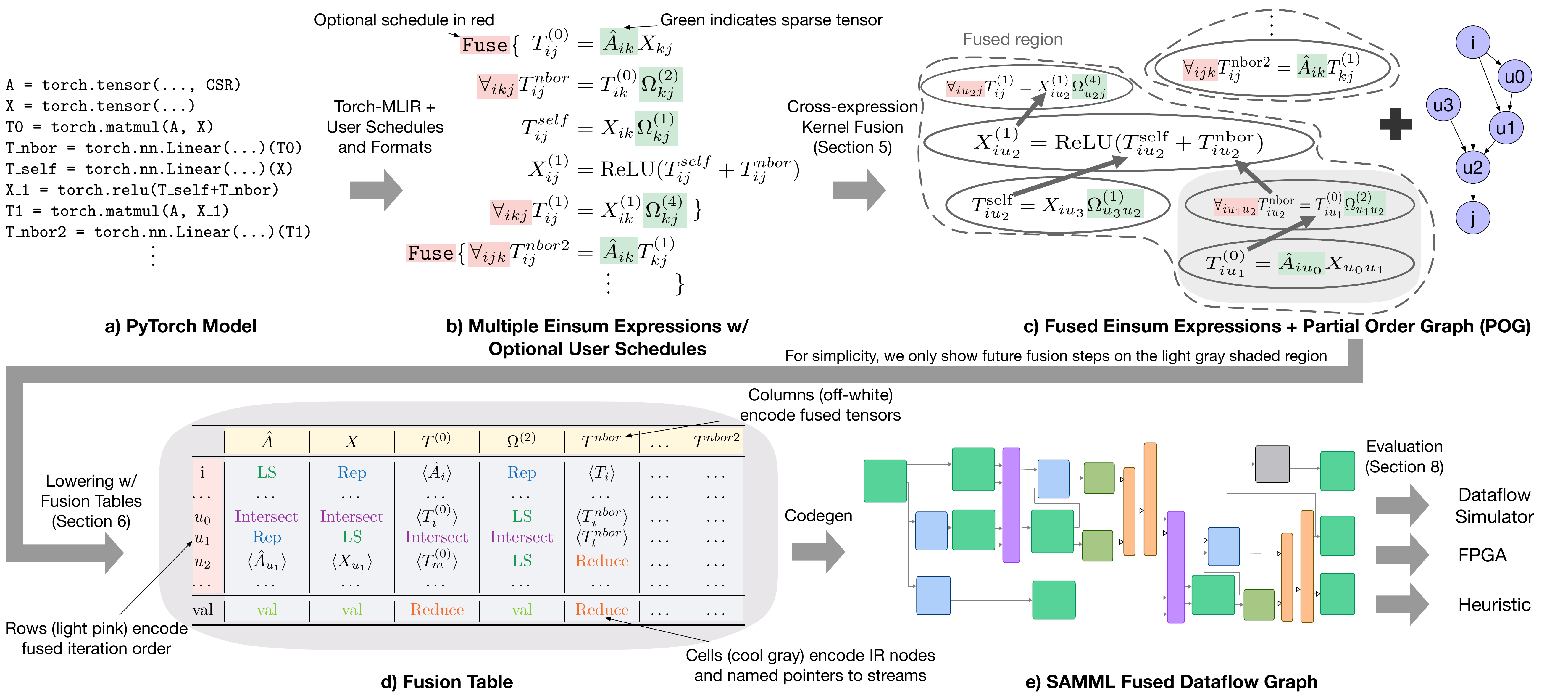}
% %\vspace{-1.0em}
% \caption{{Compilation flow of \compiler{}.} %\todo{Placeholder, update remaining results} 
\caption{\textbf{Compilation flow of \compiler{}.} (a) PyTorch model. (b) Einsum expressions with optional user schedules (red) and sparse formats (green). (c) Cross-expression fusion of \texttt{Fuse{}} regions yields fused Einsum subgraphs and a partial-order graph. (d) Fusion tables encode iteration (rows), tensors (columns), and IR nodes/streams (cells). (e) Codegen emits a SAMML fused dataflow graph; evaluation targets a dataflow simulator, FPGA, or a heuristic model (\Cref{sec:evaluation}).} 
% \todo{add in sae numbers/Normalize by unfused}
\label{fig:overview_new}
\end{figure*}

\begin{figure}[]
  \scriptsize
  \definecolor{arrowgray}{gray}{0.3}  % Define arrow color - adjust value between 0 (black) and 1 (white)
  \definecolor{arrowblue}{RGB}{48,72,168}
  % \begin{tikzpicture}[bbox]
  \begin{tikzpicture}[]
  % [
  %   execute at end picture={
  %     \draw[blue, thick]
  %       ([shift={(-3pt,-3pt)}]current bounding box.south west) rectangle
  %       ([shift={(3pt,3pt)}]current bounding box.north east);
  %   }
  % ]

    \node[draw=yellow!80!black, thick, rectangle, minimum width = 135pt, fill={rgb,255:red,253; green,242; blue,199}] (pytorch) at (0, 0) {PyTorch, TensorFlow, ...};
    \node[draw=yellow!80!black, thick, rectangle, minimum width = 135pt, fill={rgb,255:red,253; green,242; blue,199},
                                                                   anchor=north]                      (mlir) at ([yshift=-10pt]pytorch.south) {MLIR: Linalg + SparseTensor};

    % \node[draw, thick, rectangle, minimum width = 120pt, fill={rgb,255:red,174; green,190; blue,226},
    %                                                      minimum height=65pt, anchor=north]           (fuseflow_box) at ([yshift=-8pt]mlir.south) {};
    % \node[anchor=north west] at                                                                       (fuseflow_box.north west) {\shortstack{FuseFlow \\ Compiler}};
    % --- FuseFlow compiler region: indented right of sched, and right edge aligned to MLIR/PyTorch east
    % Indent from mlir.west by 10pt -> width = 140pt - 10pt = 130pt to keep east aligned.
    \def\ffindent{10pt}
    \def\ffwidth{130pt}

    % --- left vertical label anchored to the MLIR box (so it doesn't move when FuseFlow moves)
    % Anchor is north east because it is rotated
    \node[draw, thick, rectangle, anchor=north east,
          fill={rgb,255:red,219; green,231; blue,245},
          rotate=90, minimum width=70pt] (sched)
      at ([yshift=-6.5pt]mlir.south west) {Scheduling Language};

    % Fuseflow box coordinates
    \coordinate (ff_nw) at ([xshift=\ffindent]sched.south east);
    \coordinate (mlir_se) at ([xshift=-3pt]mlir.south east);
    \coordinate (ff_ne) at (sched.south east -| mlir_se);
    % \coordinate (ff_ne) at (ff_nw -| [xshift=-2pt]mlir.south east);
    % \coordinate (samml_s) at ([yshift=\ffindent]sammlir.south);
    % \coordinate (ff_sw) at (ff_nw -| samml_s);
    % \coordinate (ff_se) at (ff_ne -| samml_s);

    % \node[draw, thick, rectangle,
    %       minimum width=\ffwidth,
    %       minimum height=65pt,
    %       fill={rgb,255:red,174; green,190; blue,226},
    %       anchor=north west] (fuseflow_box)
    %   at ([xshift=\ffindent, yshift=0pt]sched.south east) {};
    \node[draw, thick, rectangle,
      fill={rgb,255:red,174; green,190; blue,226},
      minimum height=70pt,
      fit=(ff_nw) (ff_ne),
      anchor=north] (fuseflow_box) {};
    \node[anchor=north west] at (fuseflow_box.north west) {\shortstack{FuseFlow \\ Compiler}};
    
    \node[draw, thick, rectangle, anchor=north, fill={rgb,255:red,219; green,231; blue,245},]         (ff_repr) at ([yshift=-26pt]mlir.south) {\shortstack{Collection of \\ Fused CIN Representation}};
    \node[draw, thick, rectangle, anchor=north, fill={rgb,255:red,219; green,231; blue,245},]         (sammlir) at ([yshift=-15.9pt]ff_repr.south) {SAMML IR};

    % Dataflow accelerator toolchain (yellow boxes below)
    \tikzset{dfstage/.style={
      draw=yellow!80!black, thick, rectangle, align=center,
      fill={rgb,255:red,253; green,242; blue,199},
      minimum width = 60pt,
    }}

     % \node[draw, thick, rectangle, align=center,
     %  fill={rgb,255:red,253; green,242; blue,199},
     %  minimum width = 35pt, minimum height=12pt, inner sep=2pt, anchor=north]           (df_accel) at ([xshift=50pt, yshift=-8pt]sammlir.south) {{\shortstack{\scriptsize{}Dataflow \\\scriptsize{} Accelerator}}};
     
    % \node[anchor=north] at (df_accel.north) (df_accel_label) {{\shortstack{\scriptsize{}Dataflow \\\scriptsize{} Accelerator}}};
    \node[draw=yellow!80!black, thick, rectangle, minimum width = 25pt, anchor=north, fill={rgb,255:red,253; green,242; blue,199}] (fpga)
  at ([xshift=0pt, yshift=-14pt]sammlir.south) {\shortstack{FPGA Backend}};
      \node[draw=yellow!80!black, thick, rectangle, minimum width = 35pt, anchor=center, fill={rgb,255:red,253; green,242; blue,199}] (sammlsim)
  at ([xshift=-52pt, yshift=0pt]fpga.center) {\shortstack{\scriptsize{}Comal \\\scriptsize{}Simulator}};
    \coordinate (df_accel) at ([xshift=52pt, yshift=0pt]fpga.center);

    \coordinate (df_arrow_top_anchor) at ([xshift=4pt]pytorch.north east);
      
     \node[draw=yellow!80!black, thick, rectangle, minimum width = 93pt, fill={rgb,255:red,253; green,242; blue,199}, minimum height=135pt, anchor=north west]           (df_box) at ([xshift=8pt, yshift=-4pt]pytorch.north east) {};
     \node[anchor=north] at (df_box.north) (df_box_label) {\shortstack{{\underline{{Dataflow Accelerator Backend}}}}};
     
    \node[dfstage, anchor=north] (df_hwsamml) at ([xshift=0pt, yshift=-40pt]df_box_label.south)
      {\scriptsize HW-aware \\\scriptsize{} SAMML IR \cite{koul2025onyx-jssc,onyx}};  
    \node[dfstage, anchor=north] (df_cgra) at ([yshift=-18pt]df_hwsamml.south)
      {\scriptsize CGRA Dataflow Graph IR};
    \node[dfstage, anchor=north] (df_bitstream) at ([yshift=-18pt]df_cgra.south)
      {\scriptsize CGRA Bitstream~\cite{koul2023aha-tecs,koul2025onyx-jssc}};

    % % Curly arrow back
    \draw[-latex, thick, arrowgray] (df_hwsamml.east) arc[start angle=-70, end angle=180, radius=.3] node[xshift=-3pt, yshift=6.0pt, above, align=center, text=arrowgray] (end1) {};
    \coordinate (hw_cons_loc) at ([xshift=6pt, yshift=4pt]end1.north);
    \node[text=arrowgray, anchor=center] at (hw_cons_loc) (hw_cons_label) {\shortstack{\scriptsize{}HW constrai-\\nts~\cite{hsu2025thesis,koul2025onyx-jssc}}};
    \draw[-latex, thick, arrowgray] (df_box_label.south) -- node[xshift=0pt, yshift=12pt, anchor=east, align=right, text=arrowgray] {\scriptsize{}Signal\\\scriptsize{}elaboration} node[xshift=0pt, yshift=12pt, anchor=west, align=left, text=arrowgray] {\scriptsize{}\& memory \\\scriptsize{}\cite{hsu2025thesis,koul2025onyx-jssc}} (df_hwsamml);

    \draw[-latex, thick, arrowgray] (df_hwsamml) -- node[xshift=0pt, anchor=east, align=right, text=arrowgray] {\scriptsize{}CGRA \\\scriptsize{\cite{koul2023aha-tecs,koul2025onyx-jssc}} } node[xshift=0pt, anchor=west, align=left, text=arrowgray] {Mapping \\ ~} (df_cgra);
    \draw[-latex, thick, arrowgray] (df_cgra) -- node[xshift=0pt, anchor=east, align=right, text=arrowgray] {\scriptsize{}Pipelining \\ \& PnR}  node[xshift=0pt, anchor=west, align=left, text=arrowgray] {\cite{melchert2024cascade,koul2025onyx-jssc} \\ \cite{koul2023aha-tecs}} (df_bitstream);

    \draw[-latex, thick, arrowgray] (pytorch) -- node[xshift=2pt, anchor=west, align=left, text=arrowgray] {\shortstack{\scriptsize MPACT~\cite{mpact}, ...}}  node[xshift=-2pt, anchor=east, align=right, text=arrowgray] {\shortstack{Torch-MLIR~\cite{torch-mlir}, }} (mlir);
    \draw[-latex, very thick, arrowblue] (mlir) -- node[yshift=-4pt, anchor=west, align=center] {\scriptsize{}Cross-Expression\\\scriptsize{}Kernel Fusion (Sec.~\ref{sec:cross-expr-fusion})} (ff_repr);
    \draw[-latex, very thick, arrowblue] (sched) -- (fuseflow_box.west);
    \draw[-latex, very thick, arrowblue] (ff_repr) -- node[xshift=-3.1pt, anchor=east, align=center] {\scriptsize{}Lowering\\\scriptsize{}(Sec.~\ref{sec:lowering})} (sammlir);

    \coordinate (fuseflow_box_south_anchor) at (sammlir.south |- fuseflow_box.south);
    \draw[-, thick, arrowgray] (sammlir) -- (fuseflow_box_south_anchor);

    \draw[-latex, very thick, arrowblue] (fuseflow_box_south_anchor) -- (sammlsim);
    \draw[-latex, thick, arrowgray] (fuseflow_box_south_anchor) -- (fpga);
    \draw[ thick, arrowgray] (fuseflow_box_south_anchor) --  (df_accel);
    \draw[ thick, arrowgray] (df_accel) -|  (df_arrow_top_anchor);
    \draw[ thick, arrowgray] (df_arrow_top_anchor) -| (df_box.north);

    % Curly arrow back
    \draw[-, very thick, arrowblue] (sammlir.east) arc[start angle=-70, end angle=180, radius=.28] node[xshift=13pt, yshift=-7pt,above right, align=center] (end) {\scriptsize{}Optimizations\\\scriptsize{}(Sec.~\ref{sec:optimizations})};
    \draw[-latex, very thick, arrowblue] ([yshift=5pt, xshift=8pt]sammlir.north) -- ([xshift=8pt]sammlir.north);

    % logos
    \node[anchor=east] (mlirlogo) at (mlir.east) {\includegraphics[width=.25cm]{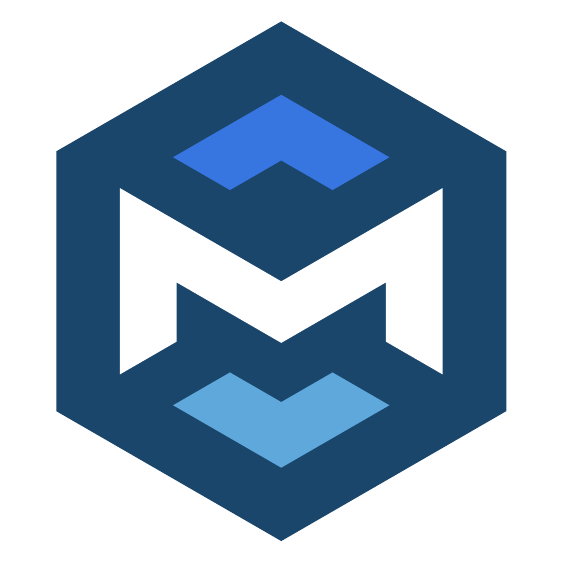}};
    \node[anchor=east] (pytorchlogo) at (pytorch.east) {\includegraphics[width=.2cm]{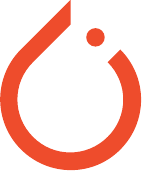}};
    \node[anchor=south east] (mlirlogo) at (fuseflow_box.south east) {\includegraphics[width=.25cm]{images/mlirlogo}};

  \end{tikzpicture}
  \caption{System overview of \compiler{}, where blue denotes new contributions.
  % \oh{Todos on this figure: add in blue arrow from SAMML to comal, consider adding in (Vitis), make yellow box outlines dark yellow color, make black arrows and text associated with those arrows dark grey, keep blue arrows "very thick" but make black arrows "thick", make "Dataflow Accelerator Backend" bold}
  }
  \label{fig:overview_system}

\end{figure}

\Cref{fig:overview_new} summarizes \compiler{}'s compilation flow from PyTorch to a fused, hardware-ready sparse dataflow graph. Models are first lowered from PyTorch~\cite{pytorch2} to MLIR (Linalg + SparseTensor dialects) using either Torch-MLIR~\cite{torch-mlir} for dense model components or MPACT~\cite{mpact} for sparse model components, 
with user-specified sparse formats and optional schedules. \revision{Since \compiler{}'s optimizations operate entirely at the MLIR Linalg + SparseTensor dialect level or lower, any frontend, which includes PyTorch through Torch-MLIR~\cite{torch-mlir}, that lowers to these dialects is supported.} This process yields a graph of Einsum expressions extended with non-algebraic operators as shown by \Cref{fig:overview_new}a to \Cref{fig:overview_new}b. This stage preserves sparsity semantics from the frontend and provides the knobs
% (formats, dataflow order)
(schedules)
that guide downstream optimization.

\subsection{\revision{Supported Sparsity Types}}
\label{sec:sparsity-types}

\revision{
\compiler{} operates on tensors whose sparse structure type is known before compilation, although the data itself does not need to be available until the generated code is executed.
The supported sparse data structure types include, in the language of the TACO data structure language~\cite{kjolstad2017taco,chou2018}, compressed data structures, uncompressed/dense data structures, coordinates, and any combination thereof in higher dimensions (e.g. dense, COO, CSR, DCSR, blocked structures, etc.).
% The supported sparse data structures include dense, COO, CSR, DCSR, n-D block structures, and combinations thereof for tensors of any dimensionality, following the TACO data structure language~\cite{kjolstad2017taco,chou2018}.
%The compiler handles any sparsity type that can be encoded in standard compressed level formats~\cite{chou2018,kjolstad2020sparse}.
This design makes \compiler{} orthogonal to the source of sparsity 
% (lossless versus lossy) 
such as from weights, activations, or inputs.}

% \textbf{Why \compiler{} is sparsity-source agnostic.}
\revision{
\compiler{}'s design is sparsity-source agnostic because its fusion algorithm (\Cref{sec:cross-expr-fusion}) and lowering (\Cref{sec:lowering}) operate on sparse tensor formats via MLIR's SparseTensor dialect, which encodes where nonzeros exist. Whether zeros arise from lossless sparsity (e.g., graph adjacency) or lossy sparsity (e.g., magnitude pruning), the resulting compressed format representation is identical. 
The system's constraints (\Cref{sec:cross-expr-fusion}) depend only on tensor mode orders and dataflow dependencies, not on sparsity provenance. Fusion tables (\Cref{sec:lowering}) organize iteration based on format structure, independent of how that structure was produced.
}

% \revision{
% The components of \compiler{}'s design that make it source agnostic are:
% \compiler{}'s fusion algorithm (\Cref{sec:cross-expr-fusion}) and lowering (\Cref{sec:lowering}) operate on sparse tensor formats via MLIR's SparseTensor dialect, which encodes where nonzeros exist. Whether zeros arise from lossless sparsity (e.g. graph adjacency), or lossy sparsity (e.g. magnitude pruning), the resulting compressed format representation is identical to \compiler{}. The system's constraints (\Cref{sec:cross-expr-fusion}) depend only on tensor mode orders and dataflow dependencies, not on sparsity provenance. Similarly, fusion tables (\Cref{sec:lowering}) organize iteration based on format structure, independent of how that structure was produced.
% }

\subsection{Compilation Flow}

\compiler{} then applies cross-expression kernel fusion to user-marked fusion regions (denoted by \texttt{Fuse}\{\} in \Cref{fig:overview_new}b), producing a fused Einsum representation 
% which we refer to as a \textit{collection of concrete index notation (CIN) expressions with fused metadata}
(\Cref{sec:cross-expr-fusion}). The fused Einsum representation includes fused components that form connected subgraph (as shown in the dashed fused region in \Cref{fig:overview_new}c) along with a partial order graph that encodes index constraints. Within the connected subgraph, the arrows indicate direct producer-consumer expression fusion.
% It inlines producer uses 
% It inlines producer results into consumers
% across kernel boundaries while building a partial order graph (\Cref{fig:overview_new}(c), right) of index constraints. 
To generate this representation, \compiler{} inlines producer results into their consumers
across kernel boundaries while building the partial order graph (\Cref{fig:overview_new}c). Partial order graph constraints are derived from the user-schedules (red) and sparse storage formats (green) (from \Cref{fig:overview_new}b).
% that encodes both dataflow order (user-marked schedule in light red in \Cref{fig:overview_new}) and sparse storage format (light green in \Cref{fig:overview_new}) constraints. 
% The result is a set of fused Einsum expressions that respect sparse storage iteration orders and optional dataflow order schedules. 
% In \Cref{fig:overview_new}(c), the arrows indicate expressions being fused. Each set of fused expression represent a connected graph as shown in the figure. 
% The dashed region in \Cref{fig:overview_new}(c) shows the resulting fused Einsum expressions, where arrows indicate direct producer-consumer expression fusion, and each fused component forms a connected subgraph.

% (\Cref{fig:overview_new}(c))

To lower the fused expressions, \compiler{} introduces a new IR called \textit{fusion tables} (\Cref{fig:overview_new}d). This representation addresses the complexity of lowering multiple fused expressions to fused dataflow graphs. The fusion table encodes: the fused iteration order based on the partial order graph in its rows (shown in \Cref{fig:overview_new}d in light pink), the fused expression which is implicit in the tensor order of its columns (shown in off-white), and IR nodes along with named pointers to intermediate streams before they are materialized in their cells (shown in cool gray).
Therefore, fusion tables represent the fused tensor computation in a tabular format before code generation, allowing for factored iteration with interleaved input iteration and computation. 
% The tabular format allows the compiler to implicitly encode the iteration order and produces named pointers to intermediate streams before they are materialized. 
Because prior work \cite{hsu2023sam, hsu2025stardust} compiles single kernels, they do not scale to face similar challenges (i.e. requiring references to temporary streams) that FuseFlow address with its fusion table IR.

\compiler{} generates SAM dataflow graphs with ML primitives, which we call SAMML. FuseFlow applies user-guided or autotuned parallelization, sparsity blocking, and dataflow-order selection, and provides a fast heuristic to estimate FLOPs/bytes for early pruning. It then executes in Comal, a cycle-accurate simulator within the open-source DAM simulation framework~\cite{dam}, or maps to FPGA backends.

\revision{
\textbf{\textit{Full Compilation Stack.}}
\Cref{fig:overview_system} shows the full compilation stack of \compiler{}. Blue components are new contributions of this work and yellow components leverage prior work for the frontend and lower-level compilation paths to hardware. Our contributions focus on algorithms for fusion and other optimizations (PyTorch $\rightarrow$ SAMML graphs), where the challenges lie for performant ML with sparse tensors on dataflow. We describe our compiler implementation and its surrounding software ecosystem in more detail in \Cref{sec:optimizations}.}

\revision{\textbf{\textit{Scheduling Language.}}}
\revision{FuseFlow exposes fusion granularity, dataflow ordering, parallelization, and a performance heuristic through explicit user schedules, enabling the design-space exploration in \Cref{sec:evaluation}. 
Users specify fusion via \emph{Fuse\{\}} regions (i.e. denoted in MLIR with functions), dataflow order via modifying Linalg affine maps, and other parameters via a command line interface. 
While this control is essential for optimal performance, future work includes autoscheduling to determine fusion schedules for common sparse ML patterns.}
\section{Cross-Expression Fusion Algorithm}
\label{sec:cross-expr-fusion}
Our solution for automatically generating fused code relies on \compiler{}'s cross-expression fusion algorithm. It fuses across distinct expressions while preserving correctness and efficiency in sparse iteration. 
Once fusion regions are scheduled, \compiler{}'s algorithm produces a collection of fused Einsum expressions and a partial order graph for each fusion region that serves as the foundation for further optimization.

\definecolor{idxcolor}{RGB}{78, 100, 250}

\begin{figure}[]
\centering
\begin{minipage}[]{.45\linewidth}
\centering
\subfloat[Unfused expressions\label{fig:initial-expr-a}]{%
  \includegraphics[width=0.9\linewidth]{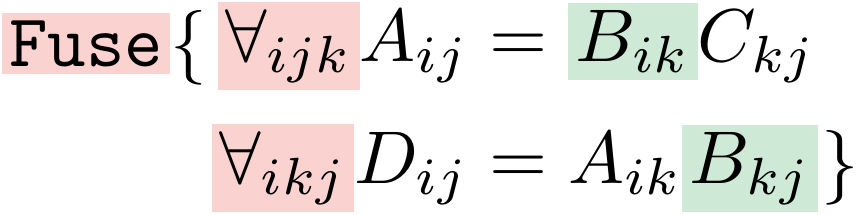}}
\par\medskip
\subfloat[Fused Einsums \label{fig:fused-einsum}]{%
  \includegraphics[width=0.9\linewidth]{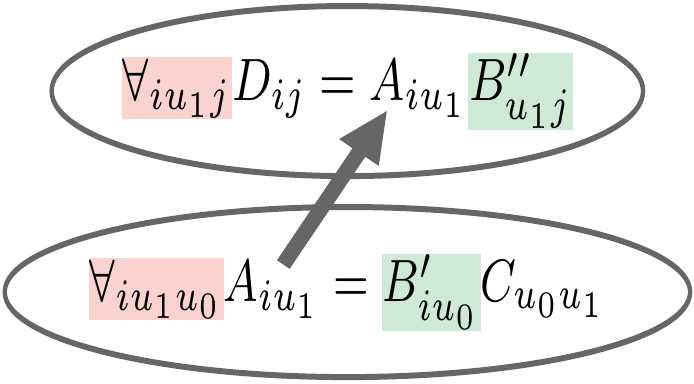}}
\end{minipage}\hfill
\begin{minipage}[]{.45\linewidth}
\centering
\subfloat[Fused Einsum w/ partial order graph\label{fig:fused-expr}]{%
\includegraphics[width=\linewidth]{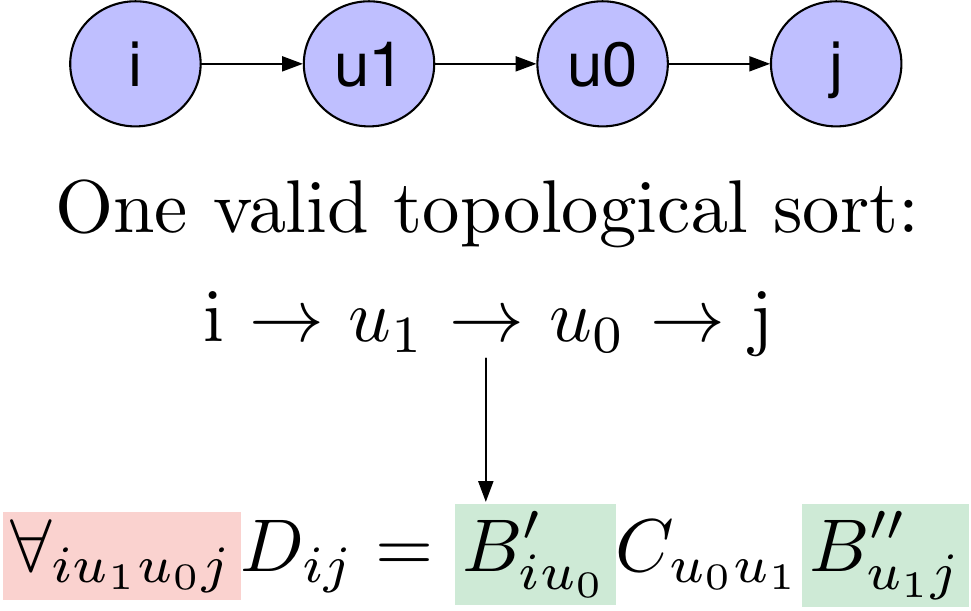}}
\end{minipage}

\caption{The input (a) with sparse matrix $B$ stored in CSR format and equivalent output representations (b) and (c) of our automated cross-expression fusion algorithm.}
\label{fig:fusion-alg}
\end{figure}

Sparsity-specific challenges arise when multiple expressions each impose their own traversal order over many sparse tensors. 
Efficient sparse \revision{tensor iteration} requires \emph{concordant traversal}, a fused iteration order compatible with each operand's native mode order (storage order)~\cite{zhang2024compilation}.
Traversing a sparse tensor against its storage format (e.g., column-wise over a CSR matrix) is \emph{discordant} and is often asymptotically worse. 
Ignoring this critical aspect of sparse \revision{tensor traversal} can lead to incorrect code~\cite{kjolstad2017taco} or suboptimal performance due to expensive indirect lookups and tensor reformatting~\cite{zhang2024compilation,kjolstad2017taco}. 

Our approach treats ordering constraints as first-class: (i) user-specified dataflow order constraints of each local expression---unspecified orders remain free---and (ii) per-tensor mode order required by storage format (e.g. CSR requires $i\rightarrow j$, denoted as $[i,j]=[0,1]$).
For example, suppose we fuse: $A_{ij}$ = $B_{ik}C_{kj}$ and $E_{ij}$ = $B_{ik}A_{kj}$, both with inner product dataflow ($i \rightarrow j \rightarrow k$). Simple index substitution conflicts with $A$'s required mode orders $[0,1]$ vs. $[1,0]$, forcing discordant traversal. 
When a tensor is used multiple times, \compiler{} treats each use as a distinct \emph{tensor view} (denoted with primes (in \Cref{fig:fused-einsum}); each view is annotated with its required mode order.

We therefore introduce a fusion algorithm that extends index substitution~\cite{debruijn1978lambda} with ordering constraints maintained in a directed graph called a partial order graph (POG).

While processing each local expression, \compiler{} inserts edges into the POG to represent both dataflow order and mode order constraints so that global consistency is maintained across the fused region. 

The POG allows \compiler{} to track index order constraints of a fused expression based on the mode order of the input sparse tensors, and final output tensor as well as the local dataflow order of each expression to be fused. This cross-expression fusion is achieved in several steps, as shown in \Cref{fig:fusion-alg}. For each expression to be fused:

    \textbf{1) Rename local index variables: } For each tensor in the expression, \compiler{} replaces all local reduction indices (those not on the left-hand side) with fresh indices (denoted as $u$-indices in \Cref{fig:fused-einsum}) and adds POG edges to enforce each sparse tensor view's mode order constraints (i.e. $i\rightarrow u_0$ based on $B'$ in \Cref{fig:fused-einsum}).
   
    \textbf{2) Build fused Einsum producer-consumer edges:} For each tensor, \compiler{} connects producer uses with their consumers, as shown by the arrows in \Cref{fig:fused-einsum}, while substituting index variables. This fused Einsum expressions representation is similar to the ideas presented by \cite{zhou2022react}.
    % \item 

    \textbf{3) Propagate order constraints:} As each producer-to-consumer edge is added, \compiler{} inserts directed edges in the POG between indices that have an outer-to-inner ordering relationship.
     
    \textbf{4) Handle multiple tensor uses:} For any tensor used multiple times, \compiler{} assigns a distinct view to each use, annotating it with the required mode order. Equivalent views are merged, where equivalence means: (i) identical mode-order sequences and (ii) equivalent index maps. If distinct views of the same tensor induce conflicting ordering constraints (detected as a cycle in the POG) and no concordant topological order exists, \compiler{} materializes a permuted copy of the tensor (a higher-order transpose) for one of the views to break the cycle.
    
By applying the above steps to every expression to be fused, we accumulate a unified index-space representation with all necessary constraints. 
Throughout this process, the POG ensures that global index ordering remains consistent with all mode orders and user-scheduled dataflow orders encountered. If the graph remains acyclic, \compiler{} performs a topological sort to produce all valid global dataflow orders that respects all constraints. 
\compiler{} can traverse the fused Einsum representation and emit a single, fully fused Einsum (\Cref{fig:fused-expr}), equivalent to the fused representation in \Cref{fig:fused-einsum}
The full algorithm can be found in \Cref{sec:full_fusion_alg}.
\section{Lowering with \Fusiontab{}s}
\label{sec:lowering}
To facilitate code generation, we introduce \textit{\fusiontab{}s}, a tabular lowering IR that memoizes intermediate streams and defers node creation. It provides named pointers to each component in the final dataflow graph, allowing for references to components that have not been created yet. 
Programming a dataflow machine differs from typical loop-based paradigms in that it relies on a spatial connection topology of operators/nodes and data rather than iterative control flow. During lowering, a dataflow compiler maps how dataflow primitives are connected to assemble the final dataflow graph.
\Fusiontab{}s capture these connections by treating each operator as a cell in a table with pointers representing data movement through nodes. \Fusiontab{}s also enable fusion across multiple expressions to target a dataflow system. We first provide details on the \fusiontab{}s (\Cref{sec:fusion_table}) and show how they are used by \compiler{} for code generation (\Cref{sec:codegen}). 

\setlength{\tabcolsep}{3pt}
\renewcommand{\arraystretch}{0.9}
\renewcommand{\cell}[1]{\ensuremath{\bm{\langle}#1\bm{\rangle}}} 

\begin{figure*}[t]
\centering
\begin{subfigure}[t]{0.20\linewidth}
\centering
\subfloat[Initial empty fusion table]
{
\begin{small}
\scriptsize
\begin{tabular}{@{}c|c|c|c@{}}
\toprule
      & $\hat{A}_{ik}$ & $X_{kj}$ & $T^{0}_{ij}$ \\
\midrule
$i$   & \textcolor{white}{} &
        \textcolor{white}{\makecell[l]{Rep(root,\\\cell{\hat{A}_{i}})}} &
        \textcolor{white}{\cell{\hat{A}_{i}}} \\ \midrule
$k$   & \textcolor{white}{LS(\cell{\hat{A}_{i}})} 
      & \textcolor{white}{LS(\cell{X_{i}})} 
      & \textcolor{white}{\cell{T^{0}_{i}}} \\[-0.02ex]
      & {\textcolor{white}{$\cap_{k}$}} 
      &  % <-- empty 3rd column
      &  % <-- empty 4th column
      \\[0.8ex]
\midrule
$j$   & \textcolor{white}{\makecell[l]{R()}} &
        \textcolor{white}{LS(\cell{X_{k}})} &
        \textcolor{white}{%
          \makecell[l]{%
            \scriptsize$\sum_{k}\bigl(\cell{\hat{A}_{\text{val}}}$\\[-0.8pt]
            \scriptsize$\cdot\cell{X_{\text{val}}}\bigr)_{\text{crd}_0}$}} \\ \midrule
val   & \textcolor{white}{V(\cell{\hat{A}_{j}})} &
        \textcolor{white}{V(\cell{X_{j}})} &
        \textcolor{white}{%
          \makecell[l]{%
            \scriptsize$\sum_{k}\bigl(\cell{\hat{A}_{\text{val}}}$\\[-0.8pt]
            \scriptsize$\cdot\cell{X_{\text{val}}}\bigr)_{\text{val}}$}} \\
\toprule
\end{tabular}
\end{small}
\label{tab:empty-table}
}
\end{subfigure}%  <-- % kills newline
\hspace{0.04\linewidth}
\begin{subfigure}[t]{0.20\linewidth}
\centering
\subfloat[Partially filled table]{
\begin{small}
\scriptsize
{
\begin{tabular}{@{}c|G|c|c@{}}
\toprule
      & $\hat{A}_{ik}$ & $X_{kj}$ & $T^{0}_{ij}$ \\
\midrule
$i$   & \textcolor{ForestGreen}{LS(root)} 
      & \textcolor{white}{\makecell[l]{Rep(root,\\\cell{\hat{A}_{i}})}} 
      & \textcolor{white}{\cell{\hat{A}_{i}}} \\
\midrule
$k$   & \textcolor{ForestGreen}{LS(\cell{\hat{A}_{i}})} 
      & \textcolor{white}{LS(\cell{X_{i}})} 
      & \textcolor{white}{\cell{T^{0}_{i}}} \\[-0.02ex]
      & {\textcolor{ltgray}{$\cap_{k}$}} 
      &  % <-- empty 3rd column
      &  % <-- empty 4th column
      \\[0.8ex]
\midrule
$j$   & \textcolor{ltgray}{\makecell[l]{Rep(\cell{\hat{A}_{k}},\\\cell{X_{j}})}} 
      & \textcolor{white}{LS(\cell{X_{k}})} 
      & \textcolor{white}{%
          \makecell[l]{%
            \scriptsize$\sum_{k}\bigl(\cell{\hat{A}_{\text{val}}}$\\[-0.8pt]
            \scriptsize$\cdot\cell{X_{\text{val}}}\bigr)_{\text{crd}_0}$}} \\
\midrule
val   & \textcolor{YellowGreen}{Val(\cell{\hat{A}_{j}})} 
      & \textcolor{white}{Val(\cell{X_{j}})} 
      & \textcolor{white}{%
          \makecell[l]{%
            \scriptsize$\sum_{k}\bigl(\cell{\hat{A}_{\text{val}}}$\\[-0.8pt]
            \scriptsize$\cdot\cell{X_{\text{val}}}\bigr)_{\text{val}}$}} \\
\toprule
\end{tabular}
}
\end{small}
\label{tab:partial-table}
}
\end{subfigure}% 
\hspace{0.058\linewidth}
\begin{subfigure}[t]{0.20\linewidth}
\centering
\subfloat[Fully constructed table]
{
\begin{small}
\scriptsize
\begin{tabular}{@{}c|G|G|G@{}}
\toprule
      & $\hat{A}_{ik}$ & $X_{kj}$ & $T^{0}_{ij}$ \\
\midrule
$i$   & \textcolor{ForestGreen}{LS(root)} &
        \textcolor{RoyalBlue}{\makecell[l]{Rep(root,\\\cell{\hat{A}_{i}})}} &
        \cell{\hat{A}_{i}} \\ \midrule
$k$   & \textcolor{ForestGreen}{LS(\cell{\hat{A}_{i}})} &
        \textcolor{ForestGreen}{LS(\cell{X_{i}})} &
        \cell{T^{0}_{i}} \\[-0.02ex]
      & \multicolumn{2}{c|}{\cellcolor{ltgray}\textcolor{Purple}{$\text{Intersect}_{k}$}} & \\[0.8ex] \midrule
$j$   & \textcolor{RoyalBlue}{\makecell[l]{Rep(\cell{\hat{A}_{k}},\\\cell{X_{j}})}} &
        \textcolor{ForestGreen}{LS(\cell{X_{k}})} &
        \textcolor{Orange}{%
          \makecell[l]{%
            $\sum_{k}\bigl(\cell{\hat{A}_{\text{val}}}$\\[-0.8pt]
            $\times\cell{X_{\text{val}}}\bigr)_{\text{crd}_0}$}} \\ \midrule
val   & \textcolor{YellowGreen}{Val(\cell{\hat{A}_{j}})} &
        \textcolor{YellowGreen}{Val(\cell{X_{j}})} &
        \textcolor{Orange}{%
          \makecell[l]{%
            \scriptsize$\sum_{k}\bigl(\cell{\hat{A}_{\text{val}}}$\\[-0.8pt]
            \scriptsize$\times\cell{X_{\text{val}}}\bigr)_{\text{val}}$}} \\
\toprule
\end{tabular}
\end{small}
\label{tab:fusion-table-matmul}
}
\end{subfigure}
% \hfill
\hspace{0.05\linewidth}
\begin{subfigure} {0.22\linewidth}
        \centering
        \subfloat[Partial SAMML graph.]{
        {
        \includegraphics[width=0.9\linewidth]{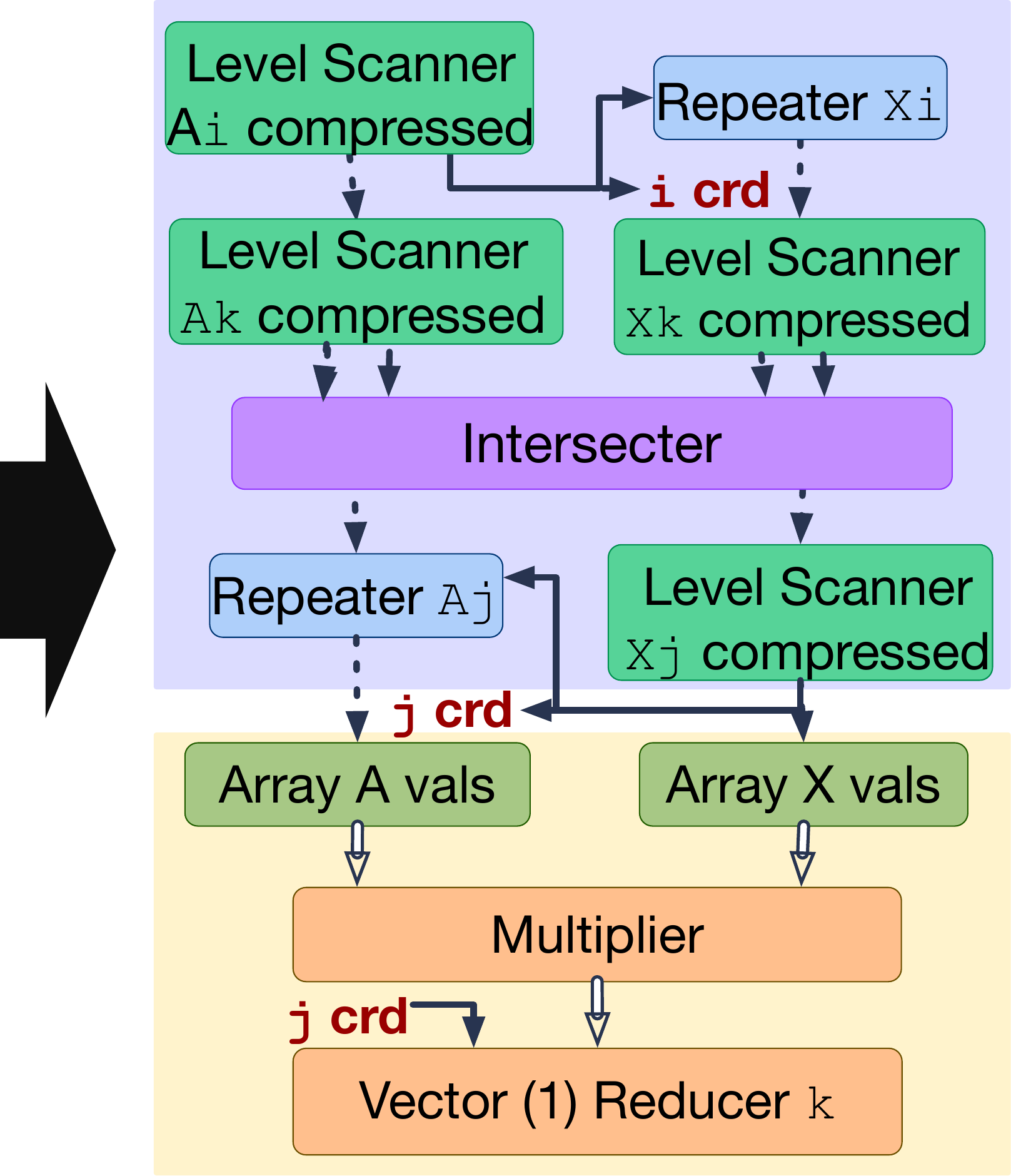}
    }
        \label{fig:sam-matmul}
    }
    \end{subfigure}

\caption{Fusion table construction for sparse matrix multiplication
    $\forall_{ikj}~T^0_{ij} = \hat{A}_{ik}X_{kj}$:
(a) an empty fusion table;
(b) a partially filled table as the compiler walks the expression DAG—highlighting how it can reference cells (nodes) not yet materialized;
(c) the fully populated fusion table;
(d) the generated dataflow graph generated from the completed table. {Colors in the \tab{} visually indicate how cell components map to nodes in \Cref{fig:sam-matmul}.} Shaded column mark which column \compiler{} has processed.}
\label{fig:fusion-table-walk}
\end{figure*}

\subsection{Fusion Table IR}
\label{sec:fusion_table}
{
{As discussed in \Cref{sec:forms_fusion}, our compiler must dynamically adjust and interleave the topology of iteration and computation pipelines. 
We use a fusion table IR to accomplish this task. A fusion table allows the compiler to defer materializing the final graph and instead work with a named, structured representation. The compiler can assign placeholders to dataflow nodes that are not yet created, enabling later pointers (references) to those future nodes by name. In essence, the fusion table provides a spatially organized plan of the fused index iteration and operations, which can be manipulated freely before committing to a final dataflow graph.

% \subsubsection{Fusion table structure}
\subsubsection*{Fusion Table Structure.} A fusion table can be thought of as a two-dimensional grid. Rows represent index variables or value results, which are ordered by fused iteration index order. For example, in \Cref{tab:empty-table} the table contains rows that represent the iteration order of the expression ($i \rightarrow k \rightarrow j$) with the last row for value computation. Columns represent tensor expressions, with each operand or intermediate result in the fused expression assigned to its own column. By reading across a given row, we see all tensors involved at that loop level. In other words, rows slice the computation by control (loop levels), while columns slice it by data (tensors). Cells occupy the intersection of a row and a column; each cell represents either an operation to perform or a pointer to another cell's operation.
A cell can be one of two types:

        {\textbf{1) Primitive cell:} creates a new dataflow node corresponding to a fundamental operation as defined in \Cref{sec:background}. Placing a primitive cell in the table is akin to planning the instantiation of that dataflow node in the SAMML graph.}
        
        \textbf{2) Reference cell: } points to an existing cell that reuses an already generated node (e.g., \cell{T^0_i} in \Cref{tab:fusion-table-matmul}) or passes through values when an index variable is not needed for the current operation (e.g. \cell{T^0_k} in \Cref{tab:fusion-table-matmul}).}

By ordering indices and operations in this structured grid, the compiler spatially captures the relationships between tensor computations and their iteration space. 
\Cref{fig:fusion-table-walk} illustrates this concept using a sparse matrix multiplication (SpMM) example.
\Cref{tab:empty-table} shows an empty fusion table for the SpMM kernel $T^0_{ij} = \sum_k A_{ik}\times X_{kj}$ with $i\rightarrow k \rightarrow j$ iteration order. \Cref{tab:partial-table} shows the table partially filled as the compiler processes the fused Einsum expressions step by step. Note that some cells are already referencing nodes (marked by angle brackets) that have not been materialized yet, indicating future connections (i.e. \cell{A_{val}} referencing \cell{A_j}). In \Cref{tab:fusion-table-matmul}, the fusion table is fully populated after handling all operations. Finally, \Cref{fig:sam-matmul} depicts the final dataflow graph generated from the completed fusion table, with the color coding showing how each cell in the table maps to a component in the final graph.

\begin{figure*}[]
\includegraphics[width=\linewidth]{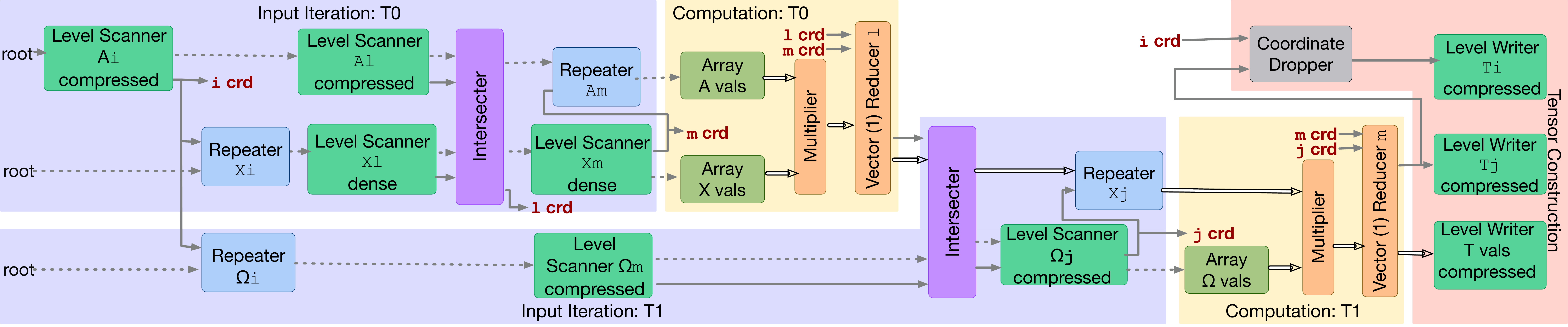}
\caption{SAMML graph for the neighborhood subcomputation $T^{nbor}_{ij} = \mathbf{\hat{A}}_{il}X_{lm}\mathbf{\Omega}^{{(2)}}_{mj}$ from GraphSAGE with $ i \rightarrow l\rightarrow m \rightarrow j$ order. See \Cref{fig:fusion-table} in \Cref{sec:complextable} for the full fusion table that generates this graph.
\label{fig:samml-neighbor}
}
\end{figure*}

\begin{figure}[]
\includegraphics[width=\linewidth]{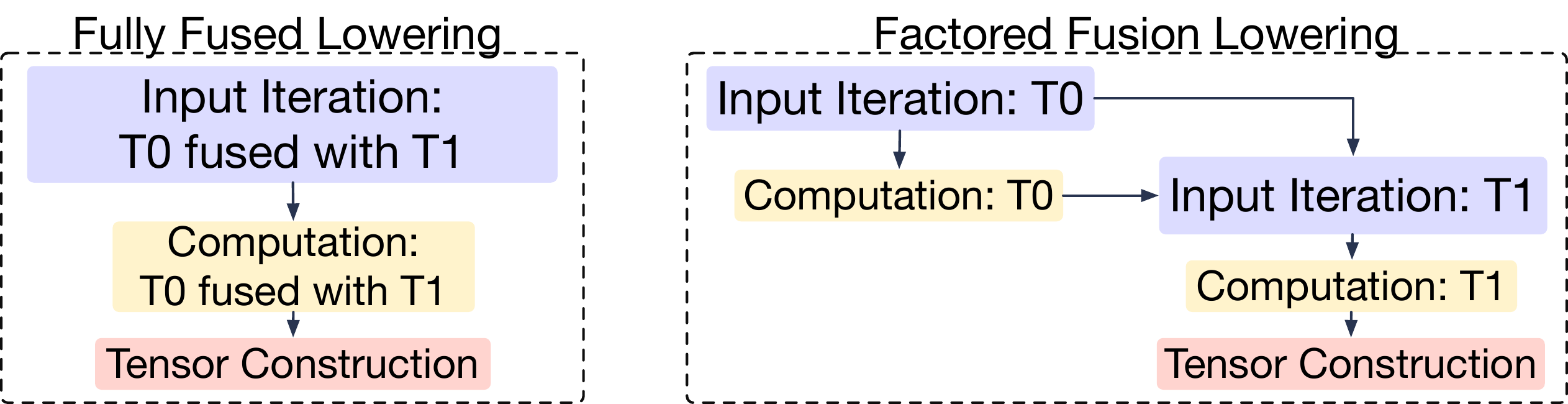}
\caption{How fully fused input iteration (\Cref{fig:fullyfused}) vs. \factored{} input iteration (\Cref{fig:binaryiterationspace}) manifests spatially in sparse dataflow graphs. Factored fusion splits the iteration sub-graphs (two blue regions), while fully-fused iteration keeps the iteration sub-graph intact (one blue region). 
\label{fig:lowering-comparison}
}
\end{figure}

\subsubsection*{Fusion Table Construction. }{To understand how fusion tables are constructed by \compiler{}, we walk through the \fusiontab{} construction for the SpMM example shown in \Cref{fig:fusion-table-walk}. \compiler{} populates \fusiontab{}s by processing each operation in the input program one by one.} \compiler{} uses several steps for each operation, as detailed below:

    \textbf{1) Insert level scanners and value nodes: } {For every input tensor view, \compiler{} assigns level scanner cells and value cells in a top-down fashion following the dataflow order. If an input tensor is not the result of a prior operation, a value cell is placed. In \Cref{tab:fusion-table-matmul}, LS cells (in dark green) for $\hat{A}_i$, $\hat{A}_k$, $\hat{X}_k$, and $\hat{X}_j$ are created, along with the corresponding value cells (labeled "Val" in light green).}
    
    \textbf{2) Insert repeat and compute nodes: } When processing intermediate tensor views, 
    \compiler{} identifies index variables missing from each input operand's tensor view and assigns Rep nodes for each of these cases. Repeat nodes' inputs include the stream being repeated and the repeat signal. The computation pipeline for the intermediate tensor view is also assigned at this point, meaning ALU and reduction nodes (if applicable) are inserted. In \Cref{tab:fusion-table-matmul}, the compute pipeline cell (in orange) is inserted for $T^0$.
    
    \textbf{3) Handle higher-order reductions: } {When a higher-order reduction is encountered, the compiler updates the relevant cells with the reduction outputs. For example, in \Cref{tab:fusion-table-matmul}, a first-order higher-order reduction (e.g., \cell{T^0_j}) is applied: it consumes a value stream and two coordinate streams. It produces a reduced value stream along with a final reduced output coordinate.}  
    
    \textbf{4) Insert stream merging nodes: } {After processing all tensors, the compiler identifies index variables shared across multiple tensor views. It then creates stream merging nodes (intersect or union) by relocating existing level scanner cells into newly created merged cells, effectively rewiring the graph before code generation. In \Cref{tab:fusion-table-matmul}, \compiler{} merges cells for $\hat{A_k}$ and $X_j$ into an intersect (shown in purple). If output coordinates coming from a higher-order reducer need merging, the corresponding reduction node's cell is moved instead.}
    This step shows how the compiler modifies the input iteration pipeline through simple cell movement.

\subsubsection*{Why fusion \revision{tables}?} 
Conventional compiler representations, such as dependency graphs or value-numbering approaches, represent computations as fixed nodes and edges \cite{hsu2023sam}. However, these approaches lack flexibility: once nodes are instantiated, it becomes challenging to reorder or restructure them dynamically without cumbersome graph transformations. 
In contrast, the fusion table is designed to be a malleable blueprint that the compiler can adjust before final code generation. This brings two key benefits:
    
    \textbf{1) Deferred graph construction for flexibility:} As described previously, \fusiontab{}s defer node creation and memoize intermediate streams, allowing for references before creation. This feature lets the compiler rewire node connections without complex graph manipulation.
    
    \textbf{2) Explicit grid layout: } 
    Rows encode fused iteration (control) and columns encode tensor views (data), with cells as operations or references. This grid makes dependencies and reuse obvious, simplifying fused graph generation and mapping cleanly to sparse dataflow hardware.
   
\subsection{Code Generation}
\label{sec:codegen}
FuseFlow generates the final dataflow graph by traversing the fusion table top-down, instantiating nodes for coordinate iteration and computation as dictated by the table structure. Starting from the output tensor value cell, FuseFlow recursively expands dependent cells upward, constructing the graph nodes and streams that correspond to the fused loops. Finally, tensor construction nodes (level writers, coordinate droppers) are added to finalize outputs. \Cref{fig:samml-neighbor} shows the final dataflow graph generated for a fused GraphSAGE kernel (see \Cref{sec:complextable} for its corresponding fusion table). The result is a hardware-efficient dataflow graph in the \factored{}-iteration style.
By design, our fusion table lowering yields a factored (not global) iteration space, interleaving input iteration and computation, as motivated in \Cref{sec:forms_fusion} and illustrated in \Cref{fig:lowering-comparison} on the right. A direct comparison of the SAM graph with global iteration space and the SAMML graph with \factored{} iteration space is shown in \Cref{sec:loweringtradeoff}.
These lowering algorithm implications are further discussed in \Cref{sec:loweringtradeoff}. 
The full lowering algorithm can be found in \Cref{sec:full_fusion_table_alg}. {Lowering \compiler{}\revision{'s} generated SAMML graph to real hardware follows prior work~\cite{onyx}, as each node lends itself to VLSI implementations. Therefore, SAMML graphs compose to represent sparse \revision{DL} on dataflow accelerators. 
\section{FuseFlow Implementation} 
\label{sec:optimizations}
We implement \compiler{} within the MLIR compiler framework.
We reimplement SAM as an MLIR dialect with a new \compiler{} MLIR compilation path as described in \Cref{sec:overview}.
% \revision{\compiler{} is front-end agnostic: our implementation targets the MLIR Linalg and SparseTensor dialects~\cite{bik}, so we support any frontend that lowers to these dialects (e.g., Torch-MLIR~\cite{torch-mlir}, MPACT~\cite{mpact}).
\compiler{} compiles these dialects to SAM graphs and lets users control fusion and dataflow ordering through a scheduling language as shown in \Cref{fig:overview_system}.
\compiler{} also includes additional optimizations, which were discussed in the context of the SAM IR but not previously present in any SAM compiler~\cite{hsu2023sam}. These optimizations---such as parallelization, sparsity blocking, dataflow ordering, and an analytical fusion heuristic---are necessary for efficient, large-scale ML.
Users can guide these optimizations through a command-line scheduling interface. 

\revision{For lower-level compilation to hardware, we leverage established infrastructure from prior work as shown in \Cref{fig:overview_system}. 
% SAMML $\rightarrow$ hardware-aware SAMML $\rightarrow$ signal elaboration and FIFO mapping $\rightarrow$ mapper $\rightarrow$ pipelining $\rightarrow$ bitstream~\cite{onyx,hsu2023sam,rucker2021capstan}.
% This stack handles hardware-aware resource allocation, including buffer sizing, compute-unit replication, and local-memory constraints.
This stack handles hardware constraints, signal elaboration, memory allocation, CGRA mapping, and pipelining with place-and-route.
Following modern compiler design principles like MLIR, separating high-level transformations from backend-specific lowering enables portability across different dataflow backends (simulator, FPGA, CGRA).}

\textit{Parallelization.} Our compiler applies vectorization and loop unrolling optimizations inspired by SAM~\cite{hsu2023sam}, concretizing them via stream parallelizer and serializer primitives. Users specify parallelization by selecting index variables and parallelization factors. The compiler partitions tensor coordinates and duplicates compute subgraphs, distributing work across parallel streams and merging results upon completion.

\textit{Sparsity Blocking. } \compiler{} efficiently targets structured sparsity (e.g., block-sparsity~\cite{zaheer2020big, dao2022flashattention}) by mapping dense blocks onto the innermost coordinates of tensors. Sparse iteration occurs at outer levels, with dense blocks streamed directly to vectorized ALUs, maintaining sparsity-driven dataflow benefits while enhancing computational density.

\textit{Dataflow Ordering. } \compiler{} enumerates valid dataflow orders that do not break fusion, enabling users or autotuning frameworks to select schedules to optimize performance~\cite{lacouture_llm_autoscheduling_2025}. Each dataflow order yields different SAMML graphs and asymptotic efficiencies.

\textit{Fusion Heuristic. } Our heuristic rapidly estimates FLOPs and memory transfers of fused programs without full simulation. Users input tensor dimensions, sparsity percentages, and intersection rates. The fusion heuristic enables lightweight analysis to quickly prune suboptimal schedules, significantly reducing the optimization search space.
\section{Evaluation} 
\label{sec:evaluation}
To evaluate the techniques presented in this paper, we use \compiler{} to compile various real-world sparse machine learning applications to our SAMML IR and simulate their end-to-end \cycleacc{} performance. 
We showcase our compiler's generality by using four different model classes. We also perform ablation studies on various key features of \compiler{} in \Cref{sec:fusion} to \Cref{sec:dataflow}. 
{\compiler{}’s general algorithmic fusion mechanism allows us to explore a vast space of fusion and dataflow schedules, unlocking speedups that were previously unattainable with existing frameworks for sparse \revision{DL} on dataflow hardware.

\subsection{Methodology}
\label{sec:methodology}

\indent \textbf{\textit{Benchmark Applications and Datasets.} }
We evaluate \compiler{} on four sparse machine learning model classes across different domains: Sparse Autoencoder (SAE)~\cite{ng2011sparse} (3 layers), Graph Convolutional Networks (GCN)~\cite{kipf2017semisupervised} (2 layers), GraphSAGE~\cite{hamilton2017graphsage} (2 layers), and GPT-3 Small (125M parameters) with BigBird attention~\cite{zaheer2020big} (sequence length of 1024). For SAE, we randomly sampled 5 images. Real world datasets \revision{(spanning 50\%-99.9\% and comprising lossless and lossy sparsity sources)} were used for each model as shown in \Cref{tab:datasets}.

\begin{table}[t]
  \centering
  \scriptsize
  \setlength{\tabcolsep}{2pt}
  \begin{tabular}{llll>{\color{black}}l}
  \toprule
    \sffamily\bfseries{Model} &
    \sffamily\bfseries{Dataset} &
    \sffamily\bfseries{MxN} &
    \sffamily\bfseries{Sparsity \%} &
    {\color{black}\sffamily\bfseries{Sparsity Source}} \\
    \midrule
    GCN/GraphSAGE & Cora~\cite{yang2016revisitingsemisupervisedlearninggraph} & 2708x1433 & {99.7\%} & ZB lossless (in) \\
    GCN/GraphSAGE & Cora\_ML~\cite{bojchevski2018deepgaussianembeddinggraphs} & 2995x2879 & {99.8\%} & ZB lossless (in) \\
    GCN/GraphSAGE & DBLP~\cite{bojchevski2018deepgaussianembeddinggraphs} & 17716x1639 & {99.6\%} & ZB lossless (in) \\
    GCN/GraphSAGE & OGB-Collab~\cite{hu2020ogb} & 235868x128 & {99.9\%} & ZB lossless (in) \\
    GCN/GraphSAGE & OGB-MAG~\cite{hu2020ogb} & 1939743x128 & {99.9\%} & ZB lossless (in) \\
    \midrule
    SAE & ImageNet~\cite{deng2009imagenet} & 224x224 & {50\%} & ZB lossy (wt) \\
    SAE & NIH-CXR~\cite{wang2017chestxray} & 1024x1024 & {50\%} & ZB lossy (wt) \\
    SAE & LUNA16~\cite{setio2017validation} & 512x512 & {50\%} & ZB lossy (wt) \\
    \midrule
    GPT-3 w/ BigBird & IMDB~\cite{bojchevski2018deepgaussianembeddinggraphs} & -- & {53.9\%-86.5\%*} & ZB lossy (mask) \\
    \bottomrule
  \end{tabular}
  \caption{
    {\revision{Datasets with sparsity levels and types. ZB = zero-based, in = input, wt = weight, mask = masked activation. {*Attention mask sparsity.}}}
    \label{tab:datasets}
  }
\end{table}

% \begin{table}[t]
%   \centering
%   \scriptsize
%   \setlength{\tabcolsep}{2pt}
%   \begin{tabular}{lllll}
%   \toprule
%     \sffamily\bfseries{Model} &
%     \sffamily\bfseries{Dataset} &
%     \sffamily\bfseries{MxN} &
%     \sffamily\bfseries{Sparsity \%} &
%     \sffamily\bfseries{Sparsity Source} \\
%     \midrule
%     GCN/GraphSAGE & Cora~\cite{yang2016revisitingsemisupervisedlearninggraph} & 2708x1433 & {99.7\%} & ZB lossless (in) \\
%     GCN/GraphSAGE & Cora\_ML~\cite{bojchevski2018deepgaussianembeddinggraphs} & 2995x2879 & {99.8\%} & ZB lossless (in) \\
%     GCN/GraphSAGE & DBLP~\cite{bojchevski2018deepgaussianembeddinggraphs} & 17716x1639 & {99.6\%} & ZB lossless (in) \\
%     GCN/GraphSAGE & OGB-Collab~\cite{hu2020ogb} & 235868x128 & {99.9\%} & ZB lossless (in) \\
%     GCN/GraphSAGE & OGB-MAG~\cite{hu2020ogb} & 1939743x128 & {99.9\%} & ZB lossless (in) \\
%     \midrule
%     SAE & ImageNet~\cite{deng2009imagenet} & 224x224 & {50\%} & ZB lossy (wt) \\
%     SAE & NIH-CXR~\cite{wang2017chestxray} & 1024x1024 & {50\%} & ZB lossy (wt) \\
%     SAE & LUNA16~\cite{setio2017validation} & 512x512 & {50}\% & ZB lossy (wt) \\
%     \midrule
%     GPT-3 w/ BigBird & IMDB~\cite{bojchevski2018deepgaussianembeddinggraphs} & -- & {53.9\%-86.5\%*} & ZB lossy (mask) \\
%     \bottomrule
%   \end{tabular}
%   \caption{
%     {\revision{Datasets with sparsity levels and types. ZB = zero-based, in = input, wt = weight, mask = masked activation. {*Attention mask sparsity.}}\oh{make the Sparsity source column blue}}
%     \label{tab:datasets}
%   }
% \end{table}

\textbf{\textit{\compiler{}  Compiler.}}
\compiler{} is implemented in MLIR within LLVM 19.1.0, with dependencies including Protobuf 28.1 and OR-Tools 9.10, and compiled using GCC 14.2.0. For all evaluated benchmarks, we select by default the first valid topological sort provided by \compiler{} (see \Cref{sec:cross-expr-fusion}).

\textbf{\textit{{Compilation Overhead.}}}
{All models compile in $<750$ms.}

\begin{figure*}[]
% {\color{blue}\fbox
\revisionfig{\includegraphics[width=\linewidth]{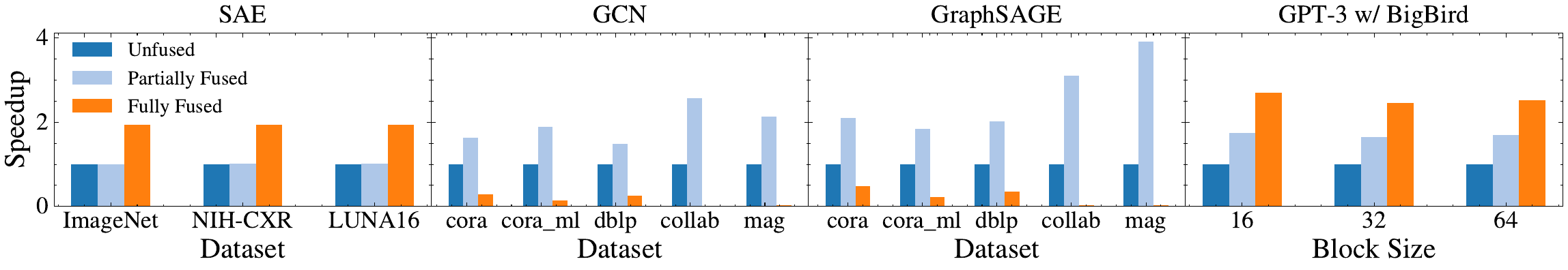}}
\caption{{The effect of fusion on dataflow performance across various models.} 
\label{fig:eval-fusion2}
}
\end{figure*}

\textbf{\textit{Simulator Framework.} }
{Our} \simulator{} {simulator} models the architectural behavior of each IR node and tracks cycles based on fully pipelined dataflow graphs, as in SAM~\cite{hsu2023sam}. 
It incorporates HBM2 memory simulation via Ramulator 2.0~\cite{luo2023ramulator2}, a cycle-accurate DRAM simulator, and provides instrumentation to estimate operations and memory accesses. \simulator{} uses the DAM simulation framework~\cite{dam} in Rust 1.87.0 with all simulation functional results verified against a dense PyTorch implementation.

\subsection{Hardware Validation}  
\label{sec:hardware_validation}
As in SAM~\cite{hsu2023sam}, our primary evaluation uses a cycle-accurate simulator. At the time of writing, no existing accelerator broadly supported end-to-end sparse ML. 
The closest, Onyx~\cite{onyx}, a coarse-grained reconfigurable array (CGRA) targeting sparse tensor algebra, is insufficient for sparse ML as it lacks support for nonlinear and masking operations.
Therefore, we validate simulator fidelity against a post-synthesis RTL simulation of a Xilinx VU9P (AWS F1) design generated from \compiler{}’s SAMML. \compiler{} lowers to Comal for simulation and to Vitis HLS for FPGA, using a minimal, proof-of-concept HLS library to instantiate streaming operators. We select kernels that fit entirely in on-chip BRAM to isolate compute, including partially fused (one-layer) GCN and GraphSAGE, fully fused GCN, and BigBird attention. Concretely, GCN (11 kernels) and GraphSAGE (13 kernels) on KarateClub~\cite{zachary1977information} and GPT-3 (17 kernels) with sequence length 64. For each model, we normalize kernel cycle counts by the best across both backends and report trend agreement via $R^2$ over kernels. We observe a strong agreement of $R^2{=}0.991$ in \Cref{fig:fpga_corr}. Fairly recently, \citet{chen2025opal} introduced Opal, a follow-on CGRA to Onyx with sparse ML support and improved dataflow orderings, which we plan to target as future work.

\begin{figure}[]
\centering
        \includegraphics[width=0.95\linewidth]{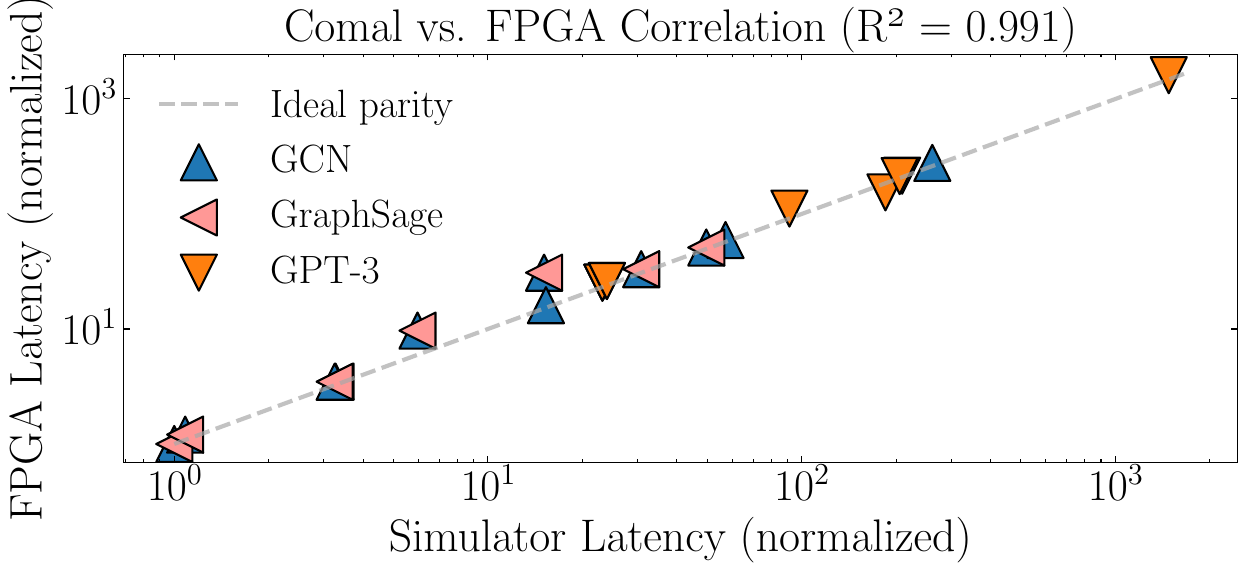}
        \caption{Latency correlation \revision{of \simulator{} simulation versus an FPGA} across kernels across various models. Colors denote models; the dashed line shows parity.}
        \label{fig:fpga_corr}
\end{figure}

% \begin{figure*}[]
%     \centering
%     \begin{minipage}[]{.7\linewidth}
%         \centering
%         \includegraphics[width=\linewidth]{images/updated_normalized.pdf}
%         \caption{
%         GCN FLOPs and memory accesses normalized to the unfused baseline across datasets. Dashed lines show operational intensity: fusion always increases operational intensity, but full fusion’s recomputation increases both FLOPs and memory.
%         \label{fig:flop_mem_access}
%         }
%     \end{minipage}
%     \hfill
    % \begin{minipage}[]{.27\linewidth}
    %     \centering
    % \scriptsize
    % \setlength{\tabcolsep}{6pt}
    % \captionof{table}{Mean absolute percent error for FLOPs and memory accesses across kernels. Experiment settings: GPT-3 (block size 16); GCN and GraphSAGE kernels on OGB-Collab.}
    % \begin{tabular}{lcc}
    %   \toprule
    %   & \multicolumn{2}{c}{Avg \% Error} \\
    %   Model class & FLOPs & Bytes \\
    %   \midrule
    %   GPT-3 & 1.8 & 5.7 \\
    %   GCN              & 2.8 & 9.6 \\
    %   GraphSAGE        & 2.6 & 11.5 \\
    %   \bottomrule
    % \end{tabular}
    % \label{tab:heuristic}
    % \end{minipage}
% \end{figure*}

\begin{table}
\begin{minipage}{0.48\columnwidth}
  \centering
  \footnotesize
  \setlength{\tabcolsep}{3pt}
  \begin{tabular}{lcc}
    \toprule
    & \multicolumn{2}{c}{Avg \% Error} \\
    Model class & FLOPs & Bytes \\
    \midrule
    GPT-3 (block=16)      & 1.8 & 5.7 \\
    GCN        & 2.8 & 9.6 \\
    GraphSAGE  & 2.6 & 11.5 \\
    \bottomrule
  \end{tabular}
  \captionof{table}{Average percent error of FLOPs and memory accesses on OGB-Collab.}
  \label{tab:heuristic}
\end{minipage}
\hfill
\begin{minipage}{0.48\columnwidth}
  \centering
  \footnotesize
  \setlength{\tabcolsep}{3pt}
  \begin{tabular}{l@{\quad}r@{\quad}r}
    \toprule
     Model &
     Unconstr. &
     Constr. \\
    \midrule
    GCN        & $2.0\cdot10^8$* & $6.3\cdot10^7$ \\
    GraphSAGE  & $3.9\cdot10^7$  & $1.1\cdot10^3$ \\
    \bottomrule
  \end{tabular}
  \captionof{table}{Number of dataflow orders with and without local constraints (*capped, estimated up to ${\sim}10^{15}$).}
  \label{tab:prune}
\end{minipage}
\end{table}

\begin{figure*}[]
    \centering
        \centering
        \includegraphics[width=0.9\linewidth]{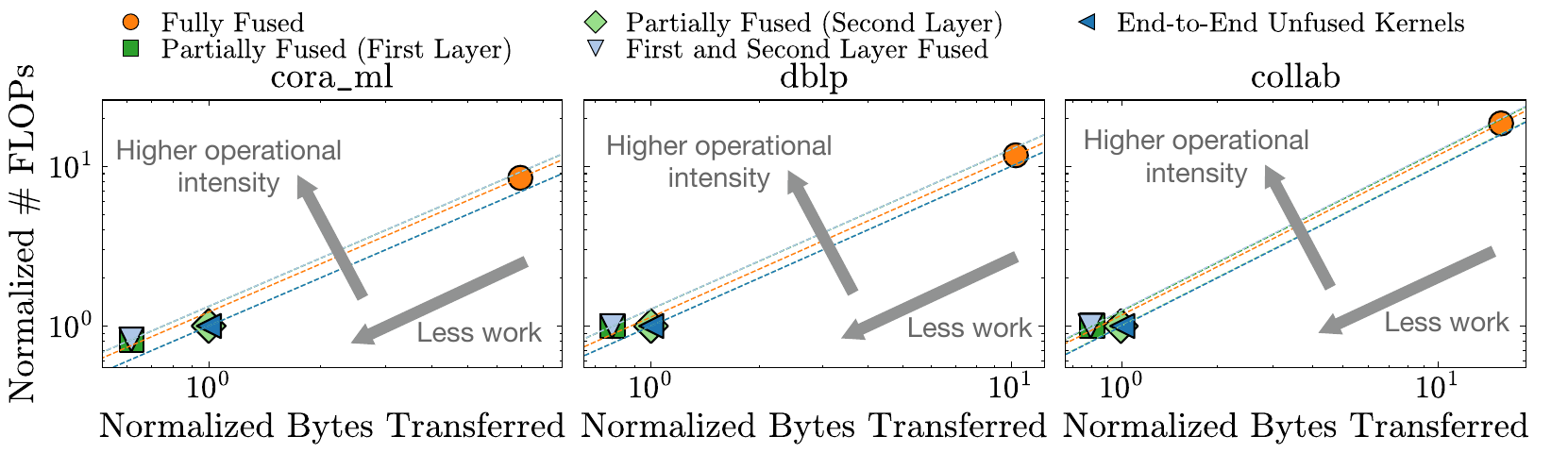}
        \caption{
        GCN FLOPs and memory accesses normalized to the unfused baseline across datasets. Dashed lines show operational intensity. Fusion increases operational intensity, but full fusion’s recomputation increases both FLOPs and memory.
        \label{fig:flop_mem_access}
        }
\end{figure*}

\subsection{Fusion}
\label{sec:fusion}
We use \compiler{} to generate fused graphs for each model comparing the performance at different fusion levels against unfused configurations. Accelerating unoptimized code is rarely useful, making fusion an essential avenue for optimization. We show that it is important to identify the right fusion granularity to ensure meaningful performance gains.
% For SAE, GCN, and GraphSAGE, full fusion marks fusion regions around the end-to-end computations, while partial fusion focuses on fusion regions within encoder/decoder or individual layers. For GPT-3 with BigBird Attention, shape operators (e.g., reshape, transpose) naturally limit fusion, dividing decoders into multiple subsets. Full fusion across decoders combines these subsets across decoder boundaries.
\paragraph{\revision{Fusion Configurations}}
\revision{We evaluate three fusion granularities: \emph{unfused} (separate operations), \emph{partially fused}, and \emph{fully fused}. For graph models and SAE, partial fusion groups operations within each layer while full fusion merges all layers. For GPT-3, reshape operations act as fusion barriers; partial fusion groups operations between reshapes within decoder blocks, while full fusion additionally merges across decoder boundaries.
See \Cref{sec:fusion_configs} in the Appendix for a visual breakdown of these exact fusion boundaries.}

As demonstrated in \Cref{fig:eval-fusion2}, GPT-3 achieves up to ${\sim}2.7\times$ improvement with full fusion. \revision{GCN and GraphSAGE experience performance degradation under full fusion due to increased computational overhead from nested matrix multiplications, so partial fusion remains more effective for these models (up to ${\sim}2.6\times$ for GCN on OGB-collab and ${\sim}3.9\times$ for GraphSAGE on OGB-mag). SAE achieves $1.94\times$ with full fusion but only ${\sim}1.01\times$ with partial fusion. Full fusion benefits from removing inter-layer materialization, but partial fusion offers limited benefit because each layer is dominated by a large sparse matrix multiplication, so fusing smaller subsequent operations provides minimal incremental gain.
% SAE results show that full fusion delivers a $1.94\times$ speedup over the unfused baseline, while partial fusion provides only ${\sim}1.01\times$ improvement. These results illustrate that optimal fusion granularity depends heavily on the model: SAE gains are primarily from removing inter-layer materialization, but each layer is still dominated by a large matrix multiplication, so fusing smaller subsequent operations (e.g., activation functions) offers limited incremental benefit. 
Models with similar patterns---large compute kernels followed by many smaller operations---will exhibit similar behavior.}

Analyzing GCN further, partially fusing the first layer significantly reduces bytes transferred compared to unfused versions, improving operational intensity (\Cref{fig:flop_mem_access}). Fully fused GCN, while having higher operational intensity, suffers from recomputation, degrading overall performance. Thus, optimal fusion must carefully balance reduced data movement against additional computation.

Finally, our heuristic effectively estimates computational and memory costs. It correctly predicts optimal fusion configurations and enables early pruning of suboptimal strategies as shown in \Cref{tab:heuristic}, which shows the average percentage errors for FLOPs and memory accesses on GPT-3 (w/ block size 16), GCN, and GraphSAGE on OGB-Collab.

\subsection{\revision{Comparison with Prior Dataflow Compilers}}
\label{sec:priorcomparison}

\revision{
We compare \compiler{} against Custard~\cite{hsu2023sam} and Stardust~\cite{hsu2025stardust} (C+S), the two prior sparse dataflow compilers for general sparse tensor algebra. As discussed in \Cref{sec:forms_fusion}, C+S only support IIF and require manual rewrites for cross-expression fusion. We evaluate on GCN with the OGB-Collab dataset~\cite{hu2020ogb} using identical simulator settings and hardware parameters across all configurations. The \emph{unfused} baseline compiles each kernel independently, materializing all intermediate tensors to memory. The \emph{C+S (rewrite)} configuration applies manual expression rewrites to C+S's inputs that force fused output code within the constraints of C+S, only support for sparse tensor computations). \emph{FuseFlow} uses our automatic cross-expression fusion techniques.
}

\revision{As shown in \Cref{tab:comp_speedup}, C+S with handwritten rewrites achieves $1.97\times$ speedup over the unfused baseline. FuseFlow achieves an additional $1.33\times$ speedup over C+S, yielding $2.63\times$ total speedup. FuseFlow's gains come from two sources: (1) automatic cross-expression fusion eliminates intermediate materializations that baseline C+S cannot fuse, and (2) factored iteration during code generation reduces coordinate processing overhead. Importantly, FuseFlow achieves this with less user effort since no manual expression rewrites as the input program to the compiler are required.}

\subsection{\revision{Sparsity Ablation Study}}
\label{sec:sparsityablation}
\revision{
To isolate sparsity's effect on fusion performance, we evaluate \compiler{} on 2-layer GCN using synthetic graphs (500 nodes, 128-dimensional features) with adjacency matrix sparsity varying from 50\% to 95\%. We test three graph structures (sparsity patterns): uniform random, power-law (scale-free networks), and block diagonal (clustered communities).
\Cref{fig:sparseablation} shows that partial fusion achieves consistent speedups that increase with sparsity, as sparser matrices reduce coordinate processing overhead. Structured patterns (power-law, block diagonal) outperform uniform random due to better locality. In contrast, full fusion incurs slowdowns when coordination overhead dominates the reduced computation. These results confirm that optimal fusion granularity depends on both sparsity level and structure. \compiler{} scales with nonzero count rather than dense dimensions, which aligns with studies from prior sparse compilers~\cite{hsu2023sam,henry2021,kjolstad2017taco,kjolstad2020sparse}.
}

\begin{figure}[ht]
\centering
        % {\color{blue}\fbox
        \revisionfig{\includegraphics[width=0.9\linewidth]{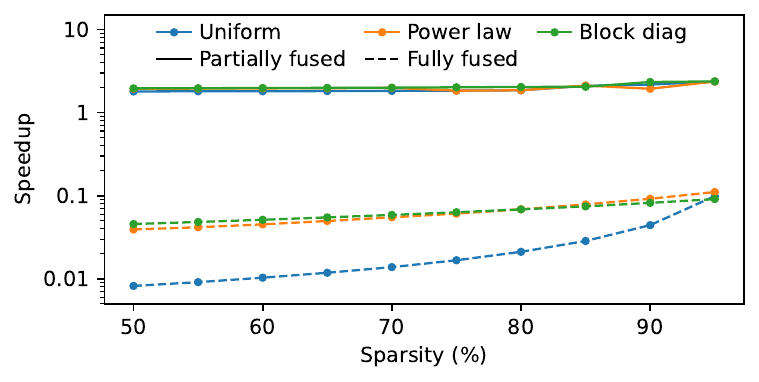}}
        \caption{\revision{Speedup over unfused baseline vs.\ adjacency matrix sparsity on a 2-layer GCN with synthetic graphs (500 nodes, 128 features) across three sparsity patterns: uniform random, power-law, and block diagonal.}}
        \label{fig:sparseablation}
\end{figure}

\subsection{Parallelization}
\label{sec:parallelization}

\begin{figure}[]
    \newcommand{\datamodelscale}{0.19\textwidth}
    \subfloat[Par factor sweep]{
        \includegraphics[width=\datamodelscale]{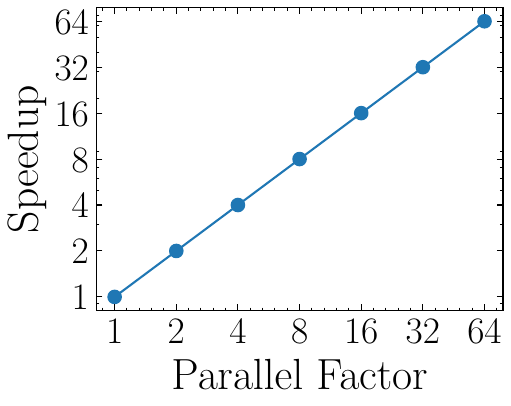}
        \label{fig:eval-par-factor}
    }
    \hspace{1em}
    \subfloat[Par location sweep]{
        \includegraphics[width=\datamodelscale]{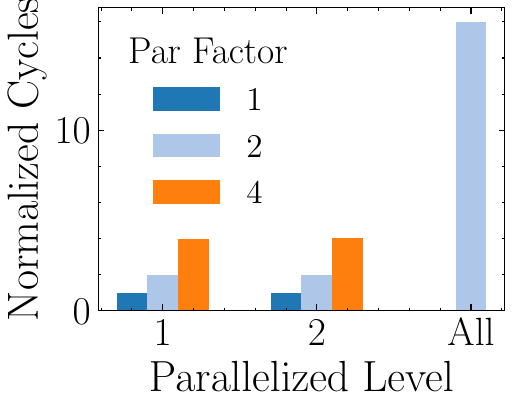}
        \label{fig:eval-par-loc}
    }
    \caption{
        The effect of parallelization factor and parallelization location for BigBird attention.
        \label{fig:eval-par} 
    }
\end{figure}

We evaluate \compiler{}'s capacity to enhance performance by generating parallel dataflow graphs. 
We first perform a sweep of parallelization factors from 1 to 64 to parallelize a single index variable in BigBird attention, as shown in \Cref{fig:eval-par-factor}. We find that \compiler{}'s generated program scales well with the amount of added parallelism. We also demonstrate \compiler{}'s ability to parallelize across different index variables, and its support for nested parallelism.
\Cref{fig:eval-par-loc} shows the impact of parallelizing two different index variables in BigBird attention, highlighting the performance effects when sweeping across various parallelization factors, as well as applying a constant factor of 4 across both levels at the same time. 
While varying parallelization location, we find that \compiler{}'s generated programs are able to obtain performance improvements relative to the parallelization factor. Parallelizing both levels at the same time by a parallel factor of 4 results in ${\sim}15.9x$ speedup.

\subsection{Block Sparse Computation}
\label{sec:blocking}

\begin{figure}[]
    \centering
    \begin{minipage}[t]{.43\linewidth}
        \centering
        \includegraphics[width=\textwidth]{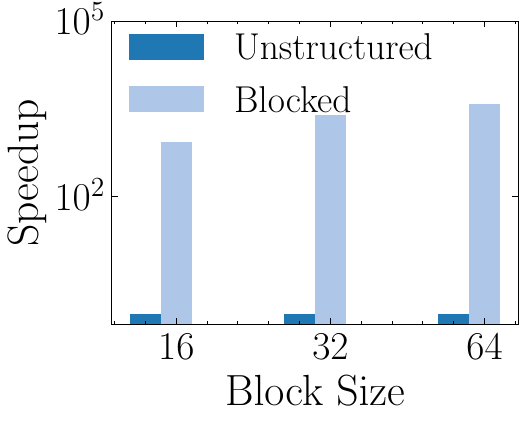}
        \caption{Performance of block sparse computation for BigBird attention.}
        \label{fig:block}
    \end{minipage}
    \hfill
    \begin{minipage}[t]{.55\linewidth}
        \centering
        \includegraphics[width=\textwidth]{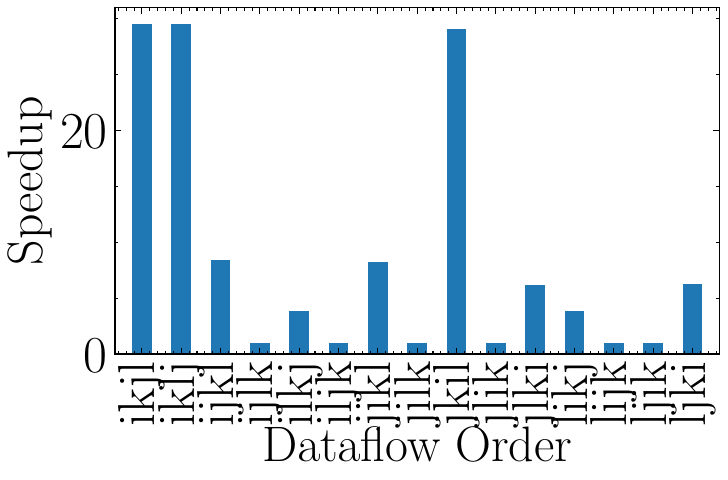}
        \caption{Dataflow order sweep for nested matmul normalized by worst dataflow.}
        \label{fig:eval-dataflow-matmul}
    \end{minipage}
\end{figure}

We evaluate \compiler{}'s sparsity blocking on the BigBird attention module for all three block configurations. 
We compare the performance of this blocked approach with our results in \Cref{fig:eval-fusion2}, which treats the tensors as unstructured sparse computation. The results in \Cref{fig:block} show that the speedup obtained is proportional to the block size.

\subsection{Dataflow Ordering}
\label{sec:dataflow}

We evaluate the impact of dataflow ordering by varying the order of nested matrix multiplication (matmul)—a core operation in GCN and GraphSAGE—using KarateClub~\cite{zachary1977information}. As shown in \Cref{fig:eval-dataflow-matmul}, suboptimal orders cause up to ${\sim}29\times$ slowdown compared to the best. Leveraging the best order thus provides an end-to-end speedup of ${\sim}29\times$ for fused GCN and GraphSAGE models.
Additionally, constraining each matmul kernel to the best dataflow order significantly reduces the design-space size by 68.5\%-99.9\%, as shown in \Cref{tab:prune}. Without these constraints, GCN alone has an impractically large number of possible dataflows (estimated up to ${\sim}10^{15}$), so we limit the search space to $2\times10^8$ configurations in \compiler{}.

% \begin{table}[]
%   \centering
%   \setlength\tabcolsep{3pt}
%   \scriptsize
%   {
%   \begin{tabular}{l@{\quad}r@{\quad}r}
%   \toprule
%     \sffamily\bfseries Model & 
%     \sffamily\bfseries Unconstr. &
%     \sffamily\bfseries Constr. \\
%     \midrule
%     GCN & $2.0\cdot10^8$* & $6.3\cdot10^7$  \\
%     GraphSAGE & $3.9\cdot10^7$ & $1.1\cdot10^3$  \\
%     \bottomrule
%   \end{tabular} 
%   }
%   \caption{
%     Number of dataflow orders with and without local dataflow constraints (*capped, estimated up to ${\sim}10^{15}$).
%   }
%     \label{tab:prune}
% \end{table}

\section{Related Work} %2 columns
\label{sec:related-work}

This paper shows how to compile sparse ML models to a sparse dataflow abstract machine. 
We review related work on sparse compilation to dataflow hardware, fusion in sparse tensor algebra frameworks, and sparse ML systems.

\subsection{Compiling Sparse Tensor Algebra to Dataflow}
Several techniques have been proposed for compiling sparse tensor algebra to dataflow hardware. Closest to our work is the Custard compiler~\cite{hsu2023sam}. Custard compiles sparse tensor algebra expressions to SAM graphs with intra-layer iteration fusion (IIF). Moreover, the compiler for the Onyx chip~\cite{onyx} maps SAM graphs to physical sparse CGRA hardware. The SAMML dataflow graphs we target are an extension of SAM graphs with additional ML primitives. Unlike Custard, our work supports a different form of IIF through factored iteration and introduces cross-expression fusion (EKF). 

An extension to the Spatial compiler~\cite{rucker2021capstan,spatial} for Capstan hardware~\cite{rucker2021capstan,plasticine} compiles computations written in parallel patterns---a loop-based declarative language---to sparse dataflow hardware. Moreover, the Stardust~\cite{hsu2025stardust} compiler can compile high-level sparse tensor algebra languages to these parallel patterns. Stardust, like Custard, only supports tensor expressions and cannot compile entire sparse ML models or provide cross-expression fusion. 

Finally, there is a class of work on compiling general-purpose C code to reconfigurable dataflow accelerators.
\citet{weng2022unifying} describe a compiler from annotated sparse loops to their SPU hardware~\cite{dadu2019spu}, and \citet{riptide} presents a co-designed compiler from general code to a CGRA. Both works support sparse loops in their general-purpose input code but do not compile from higher-level sparse languages. 

\subsection{Fusion in Sparse Tensor Frameworks} 

The TACO compiler~\cite{kjolstad2017taco} showed how to generate fused loops for sparse tensor algebra expressions on CPUs and GPUs~\cite{senanayake2020}. Later sparse tensor algebra compilers have expanded such fusion capabilities~\cite{zhou2022react} to expressions over additional data structures~\cite{ahrens2023looplets,chou2018,liu2024unisparse,sparsetir} and operations~\cite{henry2021,sundram2024recurrences,root2024shapes,kovach2023indexedstreams}. The fusion support in these compilers, however, is limited to tensor algebra expressions and CPU/GPU compilation~\cite{yadav2022spdistal}. {Other 
% specialized 
frameworks such as FusedMM~\cite{rahman2021fusedmm} and SeaStar~\cite{wu2021seastar} support sparse operator fusion but are limited to specific patterns (e.g., SDDMM+SpMM operations or GNN message-passing patterns).} 
\citet{zhou2022react} introduce techniques that identify and avoid four common redundancy types in IIF for sparse tensor algebra. Their compiler, ReACT, is most similar to ours as it introduces a representation close to our fused Einsums to and generates factored iteration code. \revision{SparseLNR~\cite{dias2022sparselnr} extends TACO with selective fusion/distribution to balance complexity and locality. Both compilers generate factored iteration code similar to ours, but they cannot fuse multiple independent expressions or generate dataflow code.}
Building upon these works, we show how to fuse across independent Einsum expressions in an ML model and how to do so when compiling to dataflow machines.

The TeAAL framework~\cite{teaal} presents a declarative language of cascaded Einsums to describe sparse tensor algebra accelerators and generates an accelerator simulator and performance model from that language. TeAAL represents multiple Einsum expressions similar to our work, but our work generates fused code across those Einsum expressions. While TeAAL is a tool for modeling dataflow accelerators, \compiler{} generates a program configuration or mapping to dataflow accelerators through the SAMML IR along with a simulation of that program.

\revision{
\subsection{Relation to Classic Loop Optimizations}
\compiler{}'s iteration-space transformations are the sparse dataflow analogues of classical loop optimizations studied extensively in prior compilers with sparse-loop optimizations~\cite{kjolstad2020sparse, bik1993,bik, venkat2015chillie} and the polyhedral compilation literature~\cite{feautrier1991dataflow,bondhugula2008practical, Strout2016SPF}.
Our intra-expression iteration fusion (IIF) corresponds to loop fusion, while dataflow order selection corresponds to loop interchange.
However, unlike traditional polyhedral compilation that operates on dense affine iteration spaces, \compiler{} fuses sparse tensor algebra operations whose iteration spaces can be thought of as polyhedra with holes~\cite{kjolstad2017taco}.
In dataflow, we operate on compressed-coordinate streams with strict ordering constraints; these streams can be viewed as a linearization of the sparse iteration space~\cite{kovach2023indexedstreams,hsu2023sam}.
Our POG encodes constraints on the loop-ordering scheduling space, analogous to dependence polyhedra, while fusion tables---similar to certain fusion information contained in schedule trees---reshape streaming dataflow primitives into a fixed iteration policy that aligns with the POG rather than rearranging imperative loop nests.
}
\subsection{Sparse ML Compilation and Frameworks}
\label{sec:polyhedral}
A few ML frameworks have been designed with support for sparse tensors and, hence, sparse ML models.  Scorch~\cite{yan2024scorch} describes several techniques needed to implement a version of the PyTorch API that supports sparse as well as dense tensors. The MLIR Linalg + SparseTensor dialects~\cite{bik} combined with the MLIR lowering from PyTorch to Linalg~\cite{mpact} also provides a sparse ML framework for CPUs. Our compilation techniques complement these frameworks with a compilation path to sparse dataflow hardware. Domain-specific libraries like PyTorch Geometric (PyG)~\cite{pytorch-geometric} and Deep Graph Library (DGL)~\cite{dgl} integrate sparse computation into specific applications, but they lack the generality needed for targeting a broader range of sparse models. 
\section{Conclusion}
\label{sec:conclusion}
{\compiler{} introduces key pieces in compiling large-scale} sparse ML models expressed in PyTorch to dataflow architectures. We believe such frameworks are essential for making productive use of these architectures. Our work opens up several avenues of future compiler work to develop further optimizations on sparse dataflow and to map from SAMML to physical RDA hardware.
\section*{Acknowledgments}

We would like to thank James Dong, Benjamin Driscoll, Chris Gyurgyik, Konstantin Hossfeld, Jungwoo Kim, Scott Kovach, Devanshu Ladsaria, Sai Gautham Ravipati, AJ Root, Alex Rucker, Nathan Sobotka, Gina Sohn, Bala Vinaithirthan, Rohan Yadav, Bobby Yan, Genghan Zhang, and Qizheng Zhang for their feedback on this paper. We would also like to thank Bo Wun Cheng and Zhouhua Xie for help on technical ideas. We would especially like to thank Shiv Sundram and Mark Horowitz for their feedback on both the work and the paper. 
This work was supported in part by the National Science Foundation under grant number 2216964, DARPA under the Machine learning and Optimization-guided Compilers for Heterogeneous Architectures (MOCHA) program (award number HR00112520038), 
and by the Naval Surface Warfare Center under Agreement
No. N00164-23-9-G057-01. 
This research was also supported in part
by the Stanford Data Analytics for What’s Next (DAWN)
Affiliate Program and the PRISM center, one of seven centers in JUMP 2.0, a Semiconductor Research Corporation (SRC) program sponsored by DARPA. Olivia Hsu was supported in part by an NSF GRFP. Any opinions, findings, and conclusions
or recommendations expressed in this material are those of
the authors and do not necessarily reflect the views of the
aforementioned funding agencies.

% use the ACM bibliography style
\bibliographystyle{ACM-Reference-Format}
\bibliography{references}

% \balance
\appendix
\pagebreak
\section{\revision{Artifact Appendix}}
\subsection{\revision{Artifact Abstract}}
\revision{This appendix describes how to set up and run the \compiler{} system, which includes programs compiled using the \compiler{} compiler and run on the \simulator{} simulator. Our artifact provides a Docker image containing all required dependencies (Python, Rust, MLIR via LLVM, protobuf, etc.) and scripts to reproduce the experimental results reported in this paper. The artifact can be executed with any x86-64 machine with Docker, Python3, Git, and Bash support, at least 64 GB of RAM, and more than 200 GB of disk space.}

\subsection{\revision{Artifact Check-List (Meta-Information)}}
\revision{
\begin{itemize}
    \item \textbf{Data set}: We use select datasets from the following sources\cite{yang2016revisitingsemisupervisedlearninggraph,bojchevski2018deepgaussianembeddinggraphs,setio2017validation,wang2017chestxray,deng2009imagenet,hu2020ogb}.
    \item \textbf{Run-time environment:} Docker, Git, Python 3, and bash need to be installed on the local machine. We recommend proficiency in bash and git.
    \item \textbf{Hardware:} Any conventional x86-64 CPU with at least 64 GB of RAM. 
    \item \textbf{Metrics:} Number of FLOPs, number of bytes transferred or accessed, search space size, latency in cycles and normalized. 
    \item \textbf{Output:} Terminal outputs, files, graphs (PDF figures).
    \item \textbf{How much disk space required (approximately)?:} Approximately  200 GB of disk space would be sufficient.
    \item \textbf{How much time is needed to prepare workflow (approximately)?:} About 5 human-minutes and 10-20 compute-minutes.
    \item \textbf{How much time is needed to complete experiments (approximately)?:}
    The total time to complete all experiments is approximately 5 human-minutes and 96 compute-hours when measured on a Google Cloud C2-standard-60 instance (60 logical threads running on a Intel Xeon Gold 6253CL Processor with 240 GB memory).
    \item \textbf{Publicly available?:} Yes, on Github at the \href{https://github.com/lrubens/fuseflow-artifact}{\color{blue}fuseflow-artifact} and on a publicly available archive \href{https://doi.org/10.6084/m9.figshare.30890834}{\color{blue}Figshare DOI}.
    \item \textbf{Code licenses (if publicly available)?:} MIT License
    \item \textbf{Workflow framework used?:} Docker
    \item \textbf{Archived (provide DOI)?}: Yes, the reserved DOI is this    \href{https://doi.org/10.6084/m9.figshare.30890834}{\color{blue}Figshare DOI}.
    % , with an anonymous link for review \href{https://figshare.com/s/6b45efb1ddf5ab29d6dd}{\color{blue}here}
\end{itemize}
}

\subsection{\revision{Description}}
\subsubsection{\revision{How to Access}} \revision{The code for this submission can be downloaded from the \href{https://github.com/lrubens/fuseflow-artifact}{\color{blue}fuseflow-artifact} repository. The repository includes a Dockerfile that can be used to build the Docker image for the full evaluation of the artifact. The artifact is also available at this reserved \href{https://doi.org/10.6084/m9.figshare.30890834}{\color{blue}Figshare DOI}.
% , with an anonymous link for review \href{https://figshare.com/s/6b45efb1ddf5ab29d6dd}{\color{blue}here}. This follows \href{https://info.figshare.com/user-guide/how-to-publish-a-dataset-at-the-same-time-as-the-associated-paper/}{\color{blue}these FigShare instructions} for anonymous artifact reviewing.
}

\subsubsection{\revision{Hardware Dependencies}} \revision{We recommend using an x86-64 machine with at least 64 GB of RAM. The more RAM available, the less compute-hours each experiment will take. The \Cref{fig:eval-fusion2} benchmark script takes in as a parameter the number of workers that dictate how many simultaneous simulations to schedule. By default we use 2, but it can be scaled up with the available memory. Running with 3 workers, it peaked at 140 GB of memory. Our compute-time estimates are calculated on a machine with 240 GB of RAM.  }

\subsubsection{\revision{Software Dependencies}} \revision{The artifact requires a machine with Docker, Git, Python 3, and bash installed. We evaluated the artifact with the following configuration Debian 6.1, Docker 20.10.24+dfsg1, Python 3.11.2, and GNU bash 5.2.15(1)-release on an Intel-based machine.}

\subsubsection{\revision{Data sets}}
\revision{
We use select datasets from the following sources\cite{yang2016revisitingsemisupervisedlearninggraph,bojchevski2018deepgaussianembeddinggraphs,setio2017validation,wang2017chestxray,deng2009imagenet,hu2020ogb}. The full set of datasets corresponding to each model can be found at~\Cref{tab:datasets}.}

\subsection{\revision{Installation}}
\revision{To install, first clone the \href{https://github.com/lrubens/fuseflow-artifact}{\color{blue}fuseflow-artifact} repository to the local machine. Then build the Docker image with the following commands (the build can take up to 20 minutes):}

\begin{lstlisting}[language=bash,basicstyle=\ttfamily\small,breaklines=true]
### Clone via HTTPS ###
$ git clone --recursive https://github.com/lrubens/fuseflow-artifact.git
$ git submodule update --init --recursive
$  docker build -t fuseflow-artifact .
\end{lstlisting}

\revision{The Docker container can be started with the following command within a bash terminal. This command will also print the container ID \texttt{CONTAINER\_ID}. }
{\small
\begin{verbatim}
$ docker run -d -it --rm fuseflow-artifact bash
\end{verbatim}
}

\revision{The container can be attached to by running:}
{\small
\begin{verbatim}
$ docker attach <CONTAINER_ID>
\end{verbatim}
}

\revision{
Once attached to the docker container, it is important not to not type \texttt{exit} in the docker terminal as this will kill the container. The proper way to exit the docker is the sequence \texttt{CTRL-p, CTRL-q}. 
}

\subsection{\revision{Experimental Workflow}}
\revision{
The experimental workflow for this artifact includes running scripts in the Docker container to run experiments and generate figures in the paper. The detailed instructions can be found in the \texttt{README.md} within the  repository.}

\subsection{\revision{Evaluation and Expected Results}}

\revision{Within the Docker container, run the following to generate all results:}

\begin{lstlisting}[
    language=bash,
    basicstyle=\ttfamily\small,
    breaklines=true,
    commentstyle=\color{green!60!black}
]
### In Docker Container ###
$ bash scripts/run_all_benchmarks.sh 
# ctrl+p ctrl+q
\end{lstlisting}

\revision{Once the experiments finish, detach the container by pressing \texttt{ctrl+p} and \texttt{ctrl+q}. To copy the experiment results and figures from the container, move outside of the \href{https://github.com/lrubens/fuseflow-artifact}{\color{blue}fuseflow-artifact} repository on the local machine and run the following commands: The \texttt{CONTAINER\_ID} is the same ID used to attach to the container. You may also retrieve the \texttt{CONTAINER\_ID} again by running \texttt{docker ps} in your terminal.
The results and figures will be copied to \texttt{fuseflow-artifact/results}.}
\begin{lstlisting}[
    language=bash,
    basicstyle=\ttfamily\small,
    breaklines=true,
    commentstyle=\color{green!60!black}
]
### In the local machine ###
# Within fuseflow-artifact/
$ bash scripts/extract_results.sh 
\end{lstlisting}

\revision{The expected results in the \texttt{fuseflow-artifact/results} directory are:}
{\small
\begin{verbatim}
fuseflow-artifact/results
|- figure12.pdf
|- figure13.pdf
|- figure14.pdf
|- figure16a.pdf
|- figure16b.pdf
|- figure17.pdf
|_ figure18.pdf
\end{verbatim}
}

\revision{
\begin{itemize}
        \item \Cref{fig:eval-fusion2}: The reproduced figure and experimental results can be found in the \texttt{fuseflow-artifact/results} folder under \texttt{figure12.pdf} and \newline \texttt{figure12\_results.json}. Verify that the results match the figure.
        \item \Cref{fig:fpga_corr}: We do not provide artifact evaluation code to reproduce the results for the hardware validation as it requires access to proprietary Xilinx FPGA tools and takes too long to synthesize the Verilog hardware.
       %  \item \Cref{fig:fpga_corr}: 
        \item \Cref{fig:flop_mem_access}: The reproduced figure and experimental results can be found in the \texttt{fuseflow-artifact/results} folder under \texttt{figure14.pdf} and \newline \texttt{figure14\_results.json}. Verify that the results match the figure.
        \item \Cref{fig:eval-par-factor}: The reproduced figure and experimental results can be found in the \texttt{fuseflow-artifact/results} folder under \texttt{figure16a.pdf} and \newline \texttt{figure16a\_results.json}. Verify that the results match the figure.
        \item \Cref{fig:eval-par-loc}: The reproduced figure and experimental results can be found in the \texttt{fuseflow-artifact/results} folder under \texttt{figure16b.pdf} and \newline
        \texttt{figure16b\_results.json}. Verify that the results match the figure.
        \item \Cref{fig:block}: The reproduced figure and experimental results can be found in the \texttt{fuseflow-artifact/results} folder under \texttt{figure17.pdf} and \newline \texttt{figure17\_results.json}. Verify that the results match the figure.
        \item \Cref{fig:eval-dataflow-matmul}: The reproduced figure and experimental results can be found in the \texttt{fuseflow-artifact/results} folder under \texttt{figure18.pdf} and \newline \texttt{figure18\_results.json}. Verify that the results match the figure.
    \end{itemize}
    }
% \pagebreak
\section{Appendix}
\subsection{Intra-expression Iteration Fusion Details}

Although we provide a diagram of intra-expression iteration fusion in \Cref{fig:fusion-forms} for dataflow, we also want to tie it to imperative loops for better understanding. We provide an example IIF fusion transformation for dense loops in \Cref{fig:iif-dense}. The transformation for sparse loops is similar but includes coiteration of sparse tensors and iteration of compressed tensor reference arrays. 

\begin{figure}[h]
    \centering
    \scriptsize
  \begin{subfigure}[b]{.40\textwidth}
    \centering
\begin{lstlisting}[basicstyle=\ttfamily\scriptsize,numbers=left,xleftmargin=20pt]
for(int i = 0; i < I; i++)
  a[i] = ...
for(int i = 0; i < I; i++)
  b[i] = ...
\end{lstlisting}
    \caption{Unfused intra-expression iteration dense loops.}
    \label{fig:unfused-dense}
  \end{subfigure}
  \begin{subfigure}[b]{.40\textwidth}
\begin{lstlisting}[basicstyle=\ttfamily\scriptsize,numbers=left,xleftmargin=20pt]
for(int i = 0; i < I; i++)
  a[i] = ...
  b[i] = ...
\end{lstlisting}
    \caption{Fused intra-expression iteration dense loops.}
    \label{fig:iif-fused-dense}
  \end{subfigure}
  \caption{Code demonstrating IIF on dense iteration spaces for dense compilers. \Cref{fig:unfused-dense} unfuses the dense iteration space for vectors a and b, while \Cref{fig:iif-fused-dense} fuses the dense iteration space for a and b. This transformation is often equivalent to loop fusion on dense loops.}
  \label{fig:iif-dense}
\end{figure}

\subsection{Full Fusion Table for GraphSAGE Example}
\label{sec:complextable}
\Cref{fig:fusion-table} shows the full fusion table for the GraphSAGE fused kernel.

\begin{figure*}[]
    \centering
    % \footnotesize
    \setlength{\tabcolsep}{2pt}
    
    \begin{tabular}{c|c|c|c|c|c}
    \toprule
         & $\hat{A}_{il}$ & $X_{lm}$ & $T^0_{im}$ & $\Omega^2_{mj}$ & $ T^{nbor}_{ij}$  \\
         \midrule
         $i$ & \textcolor{ForestGreen}{LS(root)} & \textcolor{RoyalBlue}{Rep(root, \cell{\hat{A}_{i}})} & \cell{\hat{A}_i} & \textcolor{RoyalBlue}{Rep(root, \cell{T^0_i})} & $T^0_i$ \\ 
         \midrule
         \multirow{2}{*}{$l$} & \textcolor{ForestGreen}{LS(\cell{\hat{A}_i})} & \textcolor{ForestGreen}{LS(\cell{X_i})} & \cell{T^0_i}& \cell{\Omega^{{(2)}}_i} & \cell{T^{nbor}_i} \\ 
         & \multicolumn{2}{|c|}{\textcolor{Purple}{Intersect$_l$}}  & & & \\
         \midrule
         \multirow{2}{*}{$m$} & \textcolor{RoyalBlue}{Rep(\cell{\hat{A}_l},\cell{X_m})} & \textcolor{ForestGreen}{LS(\cell{X_l})} & \textcolor{orange}{Red1$_l$} & \textcolor{ForestGreen}{LS(\cell{\Omega^{{(2)}}_l})} & \cell{T^{nbor}_l} \\ 
         & & & \multicolumn{2}{|c|}{\textcolor{Purple}{Intersect$_m$}} & \\
         \midrule
         $j$ & \cell{\hat{A}_m}& \cell{X_m} &\cell{T^0_m}[crd$_0$] & \textcolor{ForestGreen}{LS(\cell{\Omega^{{(2)}}_m})} & \textcolor{orange}{Red1$_m$[crd$_0$]} \\ 
         \midrule
        \multirow{2}{*}{val} & \multirow{2}{*}{\textcolor{YellowGreen}{Val(\cell{\hat{A}_j})}} & \multirow{2}{*}{\textcolor{YellowGreen}{Val(\cell{X_j})}} & \textcolor{RoyalBlue}{Rep(\cell{T^0_m}[val],} & \multirow{2}{*}{\textcolor{YellowGreen}{Val(\cell{\Omega^{{(2)}}_j})}} & \multirow{2}{*}{\textcolor{orange}{Red1$_m$[val]}} \\
         & & & \textcolor{RoyalBlue}{ \cell{\Omega^{{(2)}}_j})}& & \\
    \bottomrule
    \end{tabular}
    \caption{{\Fusiontab{} for our $T^{nbor}$ example from GraphSAGE where {Red1$_l$[crd0, val]  = {$\sum_l$\cell{\hat{A}_{val}} $\times$ \cell{X_{val}}}} and {Red1$_m$[crd$_0$, val] = {$\sum_m$ \cell{T^0_{val}} $\times$ \cell{\Omega^{{(2)}}_{val}}}}.} }
    \label{fig:fusion-table}
    %%\vspace{-2.5em}
\end{figure*}

\subsection{Lowering Algorithm Implications}
\label{sec:loweringtradeoff}
\begin{figure*}[t]
  \centering
  \begin{subfigure}[t]{\textwidth}
    \centering
    \includegraphics[width=\linewidth]{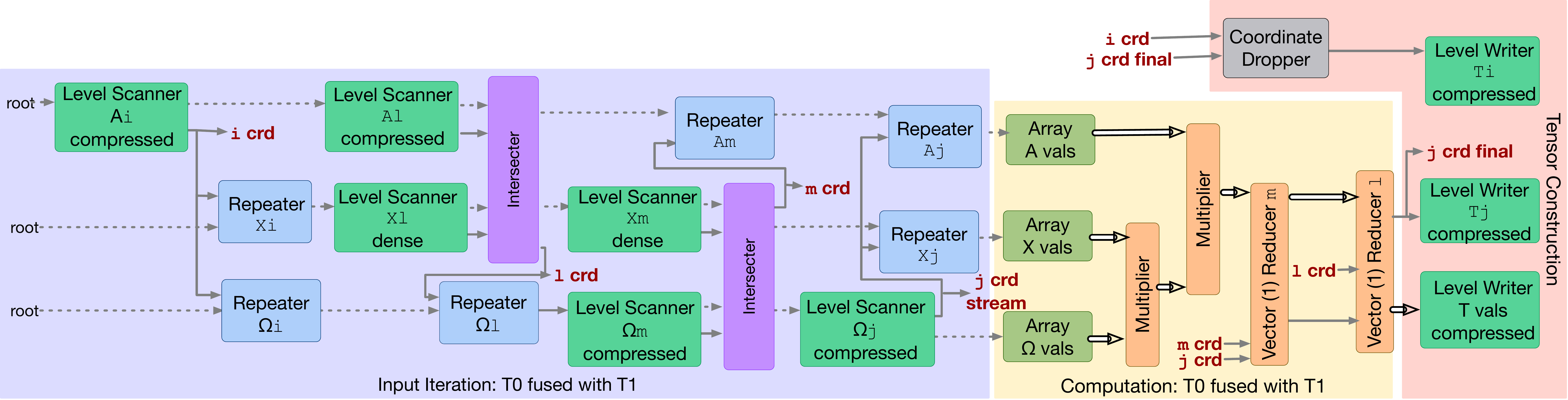}
    \caption{SAM graph with global iteration space.}
    \label{fig:sam-global}
  \end{subfigure}\hfill
  \begin{subfigure}[t]{\textwidth}
    \centering
    \includegraphics[width=\linewidth]{images/samml-graph.pdf}
    \caption{SAMML graph with factored iteration space.}
    \label{fig:samml-factored}
  \end{subfigure}
  \caption{SAM graph with global iteration space vs. SAMML graph with \factored{} iteration space.}
  % \label{fig:sam-vs-samml}
\end{figure*}
Our proposed lowering method produces dataflow graphs with computations, along with their reductions, placed in their natural positions rather than deferring them to the end. In particular, The \compiler{} compiler generates \factored{} iteration because it does not distribute multiplications across sums and does not construct a fully fused global iteration space.
{For our GraphSAGE example in \Cref{fig:samml-neighbor}, we use background color shading to help visualize this placement and distinguish interleaved regions: blue shading highlights input iteration regions, while yellow shading highlights computation regions.}  
Concretely, higher-order reducer primitives spatially appear earlier in the graph and generate coordinate streams that flow to stream joiners later in the graph. 
An abstracted version of this interleaving is shown in \Cref{fig:lowering-comparison} (right) with behavior equivalent to the factored fusion iteration space from \Cref{fig:binaryiterationspace}. On the other hand, 
\Cref{fig:lowering-comparison} (left) shows the generated SAM graphs from prior work~\cite{hsu2023sam} with its behavior equivalent to the global iteration space in \Cref{fig:fullyfused}. In this case, all computation is combined at the end, rather than interleaved. 
\Cref{fig:lowering-comparison} demonstrates how our sparse abstract machine dataflow graphs changed given the new lowering algorithm presented in this section. Concretely, for our GraphSAGE example, the SAM graph with global iteration space is shown in \Cref{fig:sam-global}, constrasting with the SAMML graph with \factored{} iteration space as shown in \Cref{fig:samml-factored}.

\subsection{Full Cross-Expression Fusion Algorithm}
\label{sec:full_fusion_alg}

\begin{algorithm*}[t]
  \caption{Cross-Expression Fusion with Ordering-Constraints}
  \label{alg:cross-expr-fusion}
  \begin{algorithmic}[1]
    \Require List of kernel expressions $\mathcal{E} = \{e_1,\dots,e_m\}$ in program order
    \Ensure Fused Einsum expressions $\mathcal{F}$ and a global partial-order graph $P=(V,E)$

    % -------------------------------------------------------------
    \Procedure{FuseExpressions}{$\mathcal{E}$}
      \Comment{Init partial order graph}
      \State $P \gets \Call{InitPOG}{}$      \Comment{nodes = index variables}
      \State $\mathcal{F} \gets [~]$         \Comment{accumulates fused kernels}

      \ForAll{expression $e \in \mathcal{E}$}           \label{l:loop-expr}
        % ---------- Step 1: rename reductions & mode-order edges ----------
        \ForAll{tensor $T$ in $e$}
          \ForAll{reduction indices $r$ of $T$}
             \State $u \gets \Call{getFreshIndexVar()}{}$       
             \State $e \gets e[r \leftarrow u]$         \Comment{substitution}
          \EndFor
          \State $E \gets E \cup \Call{ModeOrderEdges}{T}$ \Comment{add $(\cdot\!\rightarrow\!\cdot)$ edges for $T$’s format}
        \EndFor

        % ---------- Step 2: body substitution into dependent kernels ----------
        \State \Call{InlineUses}{$e,\mathcal{F}$}       \Comment{replace all uses of $e$’s outputs}

        % ---------- Step 3: propagate local data-flow order ----------
        \State $order \gets \Call{DataflowOrder}{e}$    \Comment{e.g.\ $j\!\rightarrow\!k\!\rightarrow\!i$}
        \ForAll{outer\,{$\rightarrow$}\,inner in $order$}
           \State $E \gets E \cup \{(outer,inner)\}$
        \EndFor

        % ---------- Step 4: manage tensor views & duplicates ----------
        \ForAll{tensor uses $U$ in $e$ grouped by original tensor name}
           \If{\Call{CompatibleViews}{$U$}}
              \State \Call{MergeViews}{$U$}
           \Else
              \State \Call{TagDuplicate}{$U$}           
              % \Comment{may imply higher-order transpose}
           \EndIf
        \EndFor
      \EndFor

      % ---------- Step 5: validate ordering constraints ----------
      \If{\Call{CycleDetected}{$P$}}
         \State \Call{ResolveCycles}{(P)}             \Comment{insert permutations on offending views}
      \EndIf
      \State $\pi \gets \Call{TopologicalSort}{P}$      \Comment{global concordant order}

      % ---------- Step 6: emit fused CIN kernels ----------
      \State $\mathcal{F} \gets \Call{EmitEinsum}{\pi}$    \Comment{respecting $P$ and tensor views}
      \State \Return $\mathcal{F}, P$
    \EndProcedure
  \end{algorithmic}
\end{algorithm*}
We present the cross-expression fusion algorithm as described in \Cref{sec:cross-expr-fusion} below in \Cref{alg:cross-expr-fusion}. 

\subsection{Full Fusion Table Lowering Algorithm}
\label{sec:full_fusion_table_alg}
\begin{algorithm*}[]
  \caption{Lowering a Tensor Computation Graph with Fusion Tables}
  \label{alg:fusion-table-lowering}
  \begin{algorithmic}[1]  % line numbers
    \Require Tensor-IR graph $G=(V,E)$ 
    \Ensure Staged, hardware-ready stream graph $G'$

    \Procedure{LowerGraph}{$G$}
      \State $G' \gets \Call{InitGraph}{}$
      \ForAll{tensor‐views $T$ in $V$ \textbf{(top-down)}}
        \Comment{1) Insert level scanners \& value nodes}
        \If{$T$ is \Call{InputTensor}{}}
          \State \Call{InsertLSAndVal}{$T, G'$}
        \EndIf

        \Comment{2) Insert repeat and compute nodes}
        \If{$T$ is \Call{IntermediateTensor}{}}
          \State $missing \gets$ indices absent from $T$
          \ForAll{$i \in missing$}
            \State \Call{InsertRep}{$T, i, G'$}
          \EndFor
          \State \Call{InsertComputePipeline}{$T, G'$}
        \EndIf

        \Comment{3) Handle higher-order reductions (modifies table by moving cells)}
        \If{\Call{HasHigherOrderReduction}{$T$}}
          \State \Call{LowerReduction}{$T, G'$}
        \EndIf
      \EndFor

      \Comment{4) Stream-level merging across views (modifies table by moving cells)}
      \ForAll{index-vars $i$ shared by $>1$ view}
        \State \Call{MergeStreams}{$i, G'$} \Comment{intersect / union}
      \EndFor
      \Comment{5) Emit lambda table of cell evaluators}           % step 5
      \State $\Lambda \gets \Call{EmitFusionTable}{G'}$           \Comment{$\Lambda\!:\text{cell}\mapsto\lambda$}

      \Comment{6) Trigger graph construction via output view}     % step 6
      \State $T_{\text{out}} \gets \Call{OutputTensor}{G}$
      \State \Call{Evaluate}{$\Lambda[T_{\text{out}}]$}

      \State \Return $G'$
    \EndProcedure
  \end{algorithmic}
\end{algorithm*}
We present the full fusion table lowering algorithm as described in \Cref{sec:fusion_table} below in \Cref{alg:fusion-table-lowering}.

\definecolor{layer1}{RGB}{102, 126, 234}
\definecolor{layer2}{RGB}{240, 147, 251}
\definecolor{allfused}{RGB}{250, 112, 154}

\section{\revision{Fusion Configuration Breakdown}}
\label{sec:fusion_configs}
\revision{We present the breakdown for each of the fusion configurations tested in \Cref{sec:evaluation} in \Cref{fig:fusion_configs}. Fused subset boxes align with their corresponding unfused operations to show which components are combined.}

\begin{figure*}[t]
\color{white}\fbox{\color{black}\begin{minipage}{\dimexpr\textwidth-2\fboxsep-2\fboxrule\relax}
\centering
% ============ SAE subfigure ============
\begin{subfigure}[t]{0.32\textwidth}
\centering
\begin{tikzpicture}[
    op/.style={draw, rounded corners=2pt, minimum width=1.4cm, minimum height=0.35cm,
               font=\tiny, align=center},
    fitbox/.style={draw, rounded corners=3pt, inner sep=0pt},
    coltitle/.style={font=\tiny\bfseries}
]
\def\colB{1.7cm}
\def\colC{3.4cm}
\def\vsep{1mm}

% Unfused operations
\node[op, fill=blue!15] (sae_sp1) at (0,0) {SpMM1};
\node[op, fill=green!15,  below=\vsep of sae_sp1] (sae_add1) {Add1};
\node[op, fill=purple!15, below=\vsep of sae_add1] (sae_r1) {ReLU};
\node[op, fill=cyan!15,   below=\vsep of sae_r1] (sae_sp2) {SpMM2};
\node[op, fill=orange!20, below=\vsep of sae_sp2] (sae_add2) {Add2};
\node[op, fill=red!15,    below=\vsep of sae_add2] (sae_s) {Soft};

% Column titles
\node[coltitle] at ([yshift=3mm]sae_sp1.north) {Unfused};
\node[coltitle] at ([xshift=\colB,yshift=3mm]sae_sp1.north) {Partially Fused};
\node[coltitle] at ([xshift=\colC,yshift=3mm]sae_sp1.north) {Fully Fused};

% Partially fused - use coordinates from unfused ops for proper alignment
\path ([xshift=\colB]sae_sp1.north west) coordinate (sae_k1nw);
\path ([xshift=\colB]sae_r1.south east) coordinate (sae_k1se);
\node[fitbox, fill=blue!12, fit=(sae_k1nw)(sae_k1se)] (sae_k1) {};
\node[font=\tiny, align=center] at (sae_k1.center) {Subset 1};

\path ([xshift=\colB]sae_sp2.north west) coordinate (sae_k2nw);
\path ([xshift=\colB]sae_s.south east) coordinate (sae_k2se);
\node[fitbox, fill=orange!12, fit=(sae_k2nw)(sae_k2se)] (sae_k2) {};
\node[font=\tiny, align=center] at (sae_k2.center) {Subset 2};

% Fully fused
\path ([xshift=\colC]sae_sp1.north west) coordinate (sae_kAllnw);
\path ([xshift=\colC]sae_s.south east) coordinate (sae_kAllse);
\node[fitbox, fill=blue!12, fit=(sae_kAllnw)(sae_kAllse)] (sae_kAll) {};
\node[font=\tiny, align=center] at (sae_kAll.center) {Subset 1};
\end{tikzpicture}
\caption{SAE}
\label{fig:fusion_sae}
\end{subfigure}
\hfill
% ============ GCN subfigure ============
\begin{subfigure}[t]{0.32\textwidth}
\centering
\begin{tikzpicture}[
    op/.style={draw, rounded corners=2pt, minimum width=1.4cm, minimum height=0.35cm,
               font=\tiny, align=center},
    fitbox/.style={draw, rounded corners=3pt, inner sep=0pt},
    coltitle/.style={font=\tiny\bfseries}
]
\def\colB{1.7cm}
\def\colC{3.4cm}
\def\vsep{1mm}

% Unfused operations
\node[op, fill=blue!15] (a1) at (0,0) {Adj1};
\node[op, fill=green!15,  below=\vsep of a1] (l1) {Lin mm1};
\node[op, fill=yellow!22, below=\vsep of l1] (b1) {Lin bias1};
\node[op, fill=purple!15, below=\vsep of b1] (r1) {ReLU};
\node[op, fill=cyan!15,   below=\vsep of r1] (a2) {Adj2};
\node[op, fill=lime!15,   below=\vsep of a2] (l2) {Lin mm2};
\node[op, fill=orange!20, below=\vsep of l2] (b2) {Lin bias2};
\node[op, fill=red!15,    below=\vsep of b2] (s2) {Soft};

% Column titles
\node[coltitle] at ([yshift=3mm]a1.north) {Unfused};
\node[coltitle] at ([xshift=\colB,yshift=3mm]a1.north) {Partially Fused};
\node[coltitle] at ([xshift=\colC,yshift=3mm]a1.north) {Fully Fused};

% Partially fused - use coordinates from unfused ops for proper alignment
\path ([xshift=\colB]a1.north west) coordinate (k1nw);
\path ([xshift=\colB]r1.south east) coordinate (k1se);
\node[fitbox, fill=blue!12, fit=(k1nw)(k1se)] (k1) {};
\node[font=\tiny, align=center] at (k1.center) {Subset 1};

\path ([xshift=\colB]a2.north west) coordinate (k2nw);
\path ([xshift=\colB]s2.south east) coordinate (k2se);
\node[fitbox, fill=orange!12, fit=(k2nw)(k2se)] (k2) {};
\node[font=\tiny, align=center] at (k2.center) {Subset 2};

% Fully fused
\path ([xshift=\colC]a1.north west) coordinate (kAllnw);
\path ([xshift=\colC]s2.south east) coordinate (kAllse);
\node[fitbox, fill=blue!12, fit=(kAllnw)(kAllse)] (kAll) {};
\node[font=\tiny, align=center] at (kAll.center) {Subset 1};
\end{tikzpicture}
\caption{GCN}
\label{fig:fusion_gcn}
\end{subfigure}
\hfill
% ============ GraphSAGE subfigure ============
\begin{subfigure}[t]{0.32\textwidth}
\centering
\begin{tikzpicture}[
    op/.style={draw, rounded corners=2pt, minimum width=1.4cm, minimum height=0.3cm,
               font=\tiny, align=center},
    fitbox/.style={draw, rounded corners=3pt, inner sep=0pt},
    coltitle/.style={font=\tiny\bfseries}
]
\def\colB{1.7cm}
\def\colC{3.4cm}
\def\vsep{0.3mm}

% Unfused operations
\node[op, fill=blue!15] (sa1) at (0,0) {Adj1};
\node[op, fill=green!15,  below=\vsep of sa1] (sl1a) {Lin mm1a};
\node[op, fill=yellow!22, below=\vsep of sl1a] (sb1a) {Lin bias1a};
\node[op, fill=cyan!15,   below=\vsep of sb1a] (sl1b) {Lin mm1b};
\node[op, fill=orange!20, below=\vsep of sl1b] (sb1b) {Lin bias1b};
\node[op, fill=black!8,   below=\vsep of sb1b] (sadd1) {Add};
\node[op, fill=purple!15, below=\vsep of sadd1] (sr1) {ReLU};
\node[op, fill=blue!15,   below=\vsep of sr1] (sa2) {Adj2};
\node[op, fill=green!15,  below=\vsep of sa2] (sl2a) {Lin mm2a};
\node[op, fill=yellow!22, below=\vsep of sl2a] (sb2a) {Lin bias2a};
\node[op, fill=cyan!15,   below=\vsep of sb2a] (sl2b) {Lin mm2b};
\node[op, fill=orange!20, below=\vsep of sl2b] (sb2b) {Lin bias2b};
\node[op, fill=black!8,   below=\vsep of sb2b] (sadd2) {Add};
\node[op, fill=red!15,    below=\vsep of sadd2] (ss2) {Soft};

% Column titles
\node[coltitle] at ([yshift=3mm]sa1.north) {Unfused};
\node[coltitle] at ([xshift=\colB,yshift=3mm]sa1.north) {Partially Fused};
\node[coltitle] at ([xshift=\colC,yshift=3mm]sa1.north) {Fully Fused};

% Partially fused - use coordinates from unfused ops for proper alignment
\path ([xshift=\colB]sa1.north west) coordinate (sk1nw);
\path ([xshift=\colB]sr1.south east) coordinate (sk1se);
\node[fitbox, fill=blue!12, fit=(sk1nw)(sk1se)] (sk1) {};
\node[font=\tiny, align=center] at (sk1.center) {Subset 1};

\path ([xshift=\colB]sa2.north west) coordinate (sk2nw);
\path ([xshift=\colB]ss2.south east) coordinate (sk2se);
\node[fitbox, fill=orange!12, fit=(sk2nw)(sk2se)] (sk2) {};
\node[font=\tiny, align=center] at (sk2.center) {Subset 2};

% Fully fused
\path ([xshift=\colC]sa1.north west) coordinate (skAllnw);
\path ([xshift=\colC]ss2.south east) coordinate (skAllse);
\node[fitbox, fill=blue!12, fit=(skAllnw)(skAllse)] (skAll) {};
\node[font=\tiny, align=center] at (skAll.center) {Subset 1};
\end{tikzpicture}
\caption{GraphSAGE}
\label{fig:fusion_graphsage}
\end{subfigure}

\vspace{1em}

% ============ GPT-3 subfigure ============
\begin{subfigure}[b]{\textwidth}
\centering
\begin{tikzpicture}[
    op/.style={draw, rounded corners=2pt, minimum width=0.65cm, minimum height=0.5cm,
               text depth=0pt, inner sep=1pt, font=\tiny, align=center},
    fitbox/.style={draw, rounded corners=3pt, inner sep=0pt, minimum height=0.5cm},
    coltitle/.style={font=\scriptsize\bfseries},
    reshapeop/.style={draw, rounded corners=2pt, minimum width=0.65cm, minimum height=0.5cm,
               text depth=0pt, inner sep=1pt, font=\tiny, align=center, fill=gray!30, dashed}
]

% Calculate positions for ops with smaller spacing
\def\opw{0.65cm}
\def\opsep{0.3mm}
\def\sepsep{1.5mm}

% Row 1: Unfused operations - title above
\def\urow{0.7cm}
\node[coltitle] at (8cm, \urow + 0.5cm) {Unfused};

% Decoder n ops
\node[op, fill=blue!15] (g1) at (0,\urow) {LN1};
\node[op, fill=green!15, right=\opsep of g1] (g2) {QKV\\mm};
\node[op, fill=yellow!20, right=\opsep of g2] (g3) {QKV\\bias};
\node[reshapeop, right=\opsep of g3] (g4) {Resh};
\node[op, fill=cyan!15, right=\opsep of g4] (g5) {QK\\mul};
\node[op, fill=orange!20, right=\opsep of g5] (g6) {Attn\\Mask};
\node[op, fill=lime!15, right=\opsep of g6] (g7) {Scale};
\node[op, fill=red!15, right=\opsep of g7] (g8) {Soft};
\node[op, fill=teal!15, right=\opsep of g8] (g9) {QKtV};
\node[reshapeop, right=\opsep of g9] (g10) {Resh};
\node[op, fill=blue!15, right=\opsep of g10] (g11) {Out\\mm};
\node[op, fill=green!15, right=\opsep of g11] (g12) {Out\\bias};
\node[op, fill=yellow!20, right=\opsep of g12] (g13) {Res1};
\node[op, fill=purple!15, right=\opsep of g13] (g14) {LN2};
\node[op, fill=cyan!15, right=\opsep of g14] (g15) {FFN1\\mm};
\node[op, fill=orange!20, right=\opsep of g15] (g16) {FFN1\\bias};
\node[op, fill=lime!15, right=\opsep of g16] (g17) {GeLU};
\node[op, fill=red!15, right=\opsep of g17] (g18) {FFN2\\mm};
\node[op, fill=teal!15, right=\opsep of g18] (g19) {FFN2\\bias};
\node[op, fill=pink!20, right=\opsep of g19] (g20) {Res2};

% Separator between decoder n and decoder n+1
\draw[dash pattern=on 2pt off 2pt, gray, thick] ([xshift=0.75mm]g20.east) coordinate (sep) -- ++(0, 0.7cm);
\draw[dash pattern=on 2pt off 2pt, gray, thick] ([xshift=0.75mm]g20.east) -- ++(0, -1.7cm);

% Decoder n+1 ops
\node[op, fill=blue!15, right=\sepsep of g20] (n1) {LN1};
\node[op, fill=green!15, right=\opsep of n1] (n2) {QKV\\mm};
\node[op, fill=yellow!20, right=\opsep of n2] (n3) {QKV\\bias};
\node[right=\opsep of n3, font=\tiny] (ellipsis) {...};

% Decoder labels
\node[font=\tiny, above=1mm of g1.north west, anchor=south west] {Decoder $n$};
\node[font=\tiny, above=1mm of n1.north west, anchor=south west] {Decoder $n{+}1$};

% Define common height for all boxes
\def\boxheight{0.5cm}

% Row 2: Partial fusion
\def\prow{-0.5cm}
\node[coltitle] at (8cm, \prow + 0.5cm) {Partially Fused};

\path (g1.west) ++(0,-1.2cm) coordinate (p1left);
\path (g3.east) ++(0,-1.2cm) coordinate (p1right);
\node[fitbox, fill=blue!12, minimum height=\boxheight, minimum width=0pt, fit=(p1left)(p1right)] (p1) {};
\node[font=\tiny] at (p1.center) {Subset 1};

\node[reshapeop] (p4) at (g4 |- p1) {Resh};

\path (g5.west) ++(0,-1.2cm) coordinate (p2left);
\path (g9.east) ++(0,-1.2cm) coordinate (p2right);
\node[fitbox, fill=orange!12, minimum height=\boxheight, minimum width=0pt, fit=(p2left)(p2right)] (p2) {};
\node[font=\tiny] at (p2.center) {Subset 2};

\node[reshapeop] (p10) at (g10 |- p1) {Resh};

\path (g11.west) ++(0,-1.2cm) coordinate (p3left);
\path (g20.east) ++(0,-1.2cm) coordinate (p3right);
\node[fitbox, fill=green!12, minimum height=\boxheight, minimum width=0pt, fit=(p3left)(p3right)] (p3) {};
\node[font=\tiny] at (p3.center) {Subset 3};

\path (n1.west) ++(0,-1.2cm) coordinate (p4left);
\path (n3.east) ++(0,-1.2cm) coordinate (p4right);
\node[fitbox, fill=blue!12, minimum height=\boxheight, minimum width=0pt, fit=(p4left)(p4right)] (p4box) {};
\node[font=\tiny] at (p4box.center) {Subset 1};

\node[font=\tiny] at (ellipsis |- p1) {...};

% Row 3: Fully fused
\def\frow{-1.7cm}
\node[coltitle] at (8cm, \frow + 0.5cm) {Fully Fused};

\path (g1.west) ++(0,-2.4cm) coordinate (f1left);
\path (g3.east) ++(0,-2.4cm) coordinate (f1right);
\node[fitbox, fill=blue!12, minimum height=\boxheight, minimum width=0pt, fit=(f1left)(f1right)] (f1) {};
\node[font=\tiny] at (f1.center) {Subset 1};

\node[reshapeop] (f4) at (g4 |- f1) {Resh};

\path (g5.west) ++(0,-2.4cm) coordinate (f2left);
\path (g9.east) ++(0,-2.4cm) coordinate (f2right);
\node[fitbox, fill=orange!12, minimum height=\boxheight, minimum width=0pt, fit=(f2left)(f2right)] (f2) {};
\node[font=\tiny] at (f2.center) {Subset 2};

\node[reshapeop] (f10) at (g10 |- f1) {Resh};

\path (g11.west) ++(0,-2.4cm) coordinate (f34left);
\path (n3.east) ++(0,-2.4cm) coordinate (f34right);
\node[fitbox, fill=purple!12, minimum height=\boxheight, minimum width=0pt, fit=(f34left)(f34right)] (f34) {};
\node[font=\tiny, align=center] at (f34.center) {Subset 3 + next Subset 1};

\node[font=\tiny] at (ellipsis |- f1) {...};

\end{tikzpicture}
\caption{GPT-3}
\label{fig:fusion_gpt3}
\end{subfigure}
\end{minipage}}

\caption{\revision{Fusion configurations for evaluated models. (a)--(c) show SAE, GCN, and GraphSAGE with three fusion granularities: unfused (separate kernels), partially fused (subsets per layer), and fully fused (single kernel). (d) GPT-3: reshape operations (dashed) act as fusion boundaries. Partial fusion groups operations into 3 subsets within each decoder. Fully fused merges Subset 3 of decoder $n$ with Subset 1 of decoder $n{+}1$, fusing across decoder boundaries.}}
\label{fig:fusion_configs}
\end{figure*}
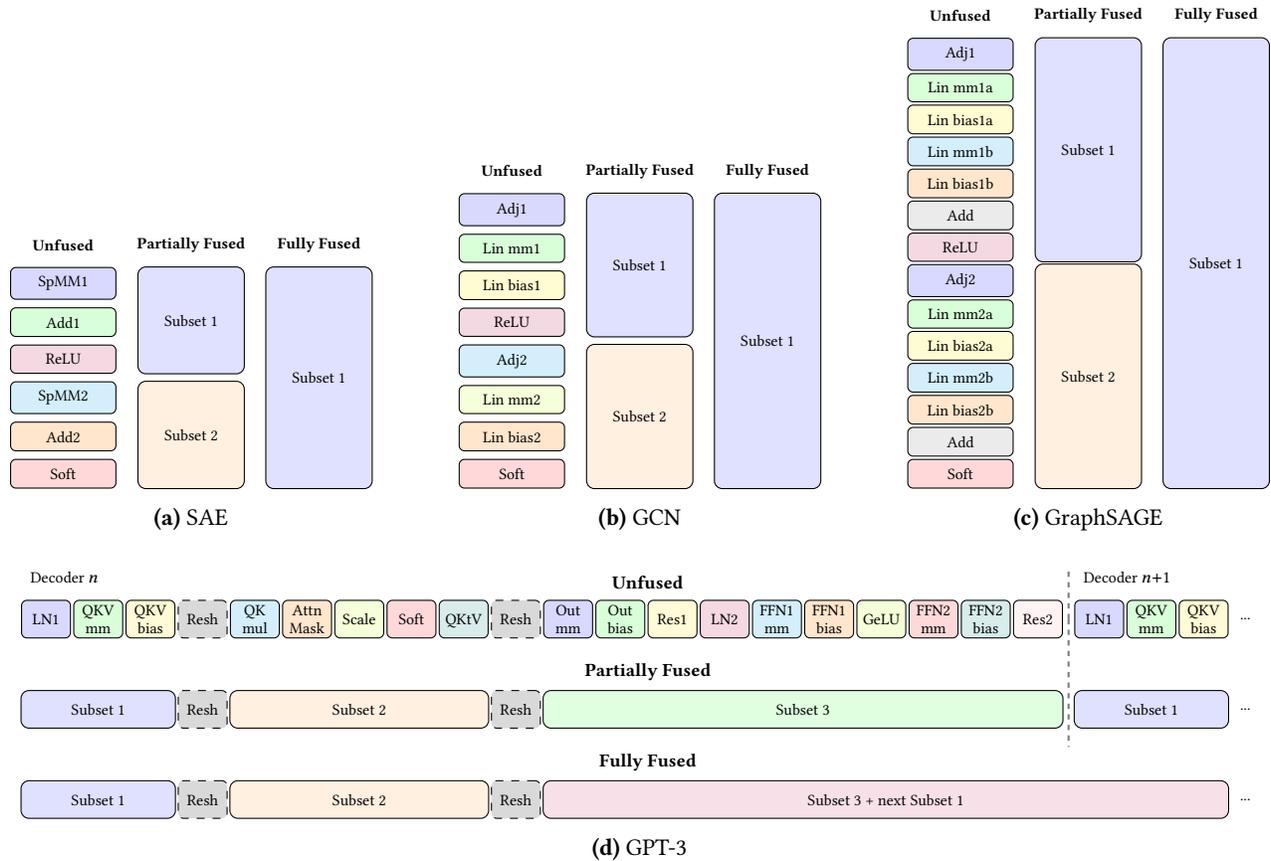

\end{document}